\documentclass{article}

\PassOptionsToPackage{numbers, compress}{natbib}
\usepackage[main, final]{neurips_2026}

\usepackage[utf8]{inputenc} %
\usepackage[T1]{fontenc}    %
\usepackage{hyperref}       %
\usepackage{url}            %
\usepackage{booktabs}       %
\usepackage{amsfonts}       %
\usepackage{nicefrac}       %
\usepackage{microtype}      %
\usepackage{xcolor}         %
\usepackage{graphicx}
\usepackage{float}
\usepackage{booktabs}
\usepackage{tikz}
\usepackage{caption}
\usepackage{subcaption}
\usepackage{enumitem}
\usepackage{pgfplots}
\usepackage[disable]{todonotes}
\usepackage{marginnote}
\usepackage{wrapfig}
\usepackage{bbm}

\usepackage{amsmath, amssymb, bm}
\usepackage{amsthm}
\theoremstyle{plain}

\newtheorem{proposition}{Proposition}
\newtheorem{lemma}{Lemma}
\newtheorem{corollary}{Corollary}
\theoremstyle{definition}
\newtheorem{definition}{Definition}
\newtheorem{remark}{Remark}

\newcommand{\E}{\mathbb{E}}
\newcommand{\R}{\mathbb{R}}

\newcommand{\inner}[2]{\left\langle #1, #2 \right\rangle}

\newcommand{\diag}{\mathrm{diag}}

\newcommand{\Cov}{\operatorname{Cov}}

\newcommand{\vecop}{\operatorname{vec}}

\newcommand{\ones}{\mathbf 1}
\newcommand{\ind}{\mathbf{1}}
\newcommand{\indep}{\mathrel{\perp\!\!\!\perp}}
\newcommand{\tildA}{\widetilde A}
\newcommand{\tildD}{\widetilde D}
\newcommand{\cR}{\mathcal R}

\newcommand{\mse}{\operatorname{MSE}}

\newcounter{result}
\newcommand{\result}[1]{%
  \par\vspace{2pt}\noindent
  \refstepcounter{result}%
  \textbf{Result \theresult: #1.}\quad\ignorespaces
}

\theoremstyle{definition}

\theoremstyle{remark}

\makeatletter
\newcommand{\authornote}[2]{\@ifundefined{@captype}{\todo[#1]{#2}}{{\let\marginpar\marginnote\todo[#1]{#2}}}}
\makeatother

\AtBeginDocument{\setlength{\marginparwidth}{84pt}\setlength{\marginparsep}{8pt}}
\newcommand{\showminisection}{0}

\newcommand{\minisection}[1]{%
  \ifnum\showminisection=1%
    \textcolor{purple}{\textit{#1.}}\quad%
  \fi%
}
\title{Bias, Measurement Error, and Double-Dipping: When Can GNN Convolutions Help Brain Connectome Prediction?}

\author{%
  Tommaso Castellani \\
  Columbia University, USA \\
  \texttt{tc3527@columbia.edu} \\
  \And
  Jiaqi Li \\
  Rice University, USA \\
  \texttt{jiaqi.li@rice.edu} \\
  \And
  Muriah D.~Wheelock \\
  Washington University in St.~Louis, USA \\
  \texttt{mdwheelock@wustl.edu} \\
  \And
  Rezwana Rosen Razzaque \\
  Washington University in St.~Louis, USA \\
  \texttt{r.razzaque@wustl.edu} \\
  \And
  Helmut Laufs \\
  Christian-Albrechts University Kiel, Germany \\
  \texttt{h.laufs@neurologie.uni-kiel.de} \\
  \And
  Hong Chen \\
  Washington University in St.~Louis, USA \\
  \texttt{chen.h1@wustl.edu} \\
  \And
  Claire Donnat \\
  University of Chicago, USA \\
  \texttt{cdonnat@uchicago.edu} \\
}

\NewDocumentEnvironment{newtext}{}{}{}

\begin{document}

\maketitle

\begin{abstract}
Graph Neural Networks (GNNs) are an important tool for fMRI prediction, where empirical functional connectivity matrices serve both as features and for graph construction. However, recent studies and our own experiments show that these methods often fail to outperform simpler graph-agnostic baselines. We propose a statistical explanation for this failure by formulating connectome prediction as an errors-in-variables (EIV) problem: the true connectome is latent, while the observed covariance matrix is a noisy finite-time-series estimate. In a linear asymptotic setting, we decompose the effect of graph convolution into three mechanisms: smoothing-induced bias and variance reduction, graph estimation error, and a double-dipping effect arising when the same noisy matrix determines both graph and features. Under a latent community model adapted to neuroscience, we identify a narrow favorable regime, set by the brain community structure, the predictive signal, the graph construction and the recording length, in which message passing can improve prediction. Simulations validate this decomposition, while real-data experiments illustrate that connectome GNN pipelines often fall outside this regime. Our results provide a first step towards  explaining when graph convolutions can help brain connectome prediction, and why they often do not.\footnote{Code is available at \url{https://github.com/tommicaste/B-ME-DD}.}

\end{abstract}

\section{Introduction}\label{sec:introduction}

Functional brain network analysis has emerged as an important approach both for diagnosing neurological disorders and for predicting cognitive characteristics.
In this setting, the data typically consist of functional magnetic resonance imaging (fMRI) recordings, captured as time-series matrices $Z_i \in \mathbb{R}^{T \times p}$; each matrix tracks how brain activity unfolds over $T$ time points across $p$ predefined regions of interest (ROIs) for a given patient.
In this context, the primary research question typically concerns how interactions among brain regions, often measured as covariances or correlations between ROIs and often referred to as ``the connectome'', relate to a subject-specific outcome,  represented by $y_i$.
This framework corresponds to positing an underlying model $f^\star$ that links the connectome to the outcome:
    $f^\star: \Sigma^{(i)} \to y_i$.

Since brain functions are hypothesized to stem from a complex network of interconnected regions~\cite{Bullmore2009}, graph-based methods are increasingly adopted for modeling the brain connectome. Following the success of GNNs in other applications, algorithms based on GNN message-passing schemes have become prevalent in the field~\cite{Cui_2023, said2024neurographbenchmarksgraphmachine}, frequently serving as the backbone of state-of-the-art pipelines with an increasing range of variants created and customized for neuroscience applications. 

\begin{wrapfigure}[28]{r}{0.5\textwidth}
    \centering
    \includegraphics[width=0.5\textwidth]{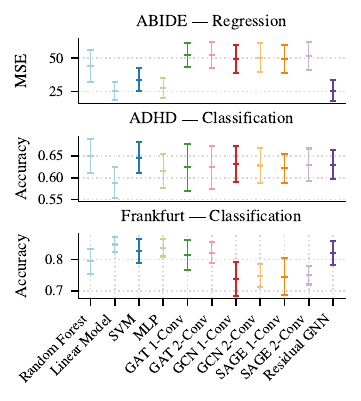}
    \captionsetup{justification=justified, singlelinecheck=false}
    \caption{Performance across regression and classification tasks on 3 datasets. Graph-based methods fail to outperform graph-agnostic baselines, both linear (Linear Model) and non-linear (Random Forest, SVM, MLP). Mean $\pm$ 1 SD over 10 seeds. See Appendix~\ref{app:real_data} for details.}
\label{fig:realdata}
    \label{fig:abide_adhd_perf}
\end{wrapfigure}Despite the popularity of graph-based methods, a growing body of evidence suggests that message passing does not actually help in analyzing the human brain connectome. In Figure~\ref{fig:abide_adhd_perf}, we show that the performance of standard GNNs (GraphSAGE \cite{hamilton2018inductiverepresentationlearninglarge}, GCN~\cite{kipf2017semisupervisedclassificationgraphconvolutional} and GAT~\cite{velickovic2018graphattentionnetworks}) and a popular brain connectomics graph methodology (Residual GNN~\cite{said2024neurographbenchmarksgraphmachine}) does not improve upon simpler (graph-agnostic) methods on three real-world datasets. This echoes the findings of~\cite{han2025rethinkingfunctionalbrainconnectome}, who benchmarked graph deep learning models on large-scale functional connectome datasets and found that their message aggregation brings no gain over graph-agnostic baselines; we refer to their work for an exhaustive empirical demonstration of the problem.
Similarly,~\cite{yang2025do} showed that basic classifiers using functional connectivity matrices consistently outperform message passing architectures. Finally, \cite{lararangel2025limitsapplyinggraphtransformers} showed that the message aggregation mechanisms characteristic of GNNs often have a detrimental effect on prediction performance from functional brain networks. From the broader GNNs landscape, the literature has focused on understanding the pitfalls of message passing in node-level tasks such as semi-supervised classification, with particular emphasis on the oversmoothing phenomenon~\citep{li2018deeperinsightsgraphconvolutional, keriven2022littlemuchtheoreticalanalysis, chen2025understandingeffectgcnconvolutions, rusch2023surveyoversmoothinggraphneural}. However, these critiques are either predominantly empirical or in a non-applicable setting; hence our main question of interest:
\begin{center}
\emph{Why do graph message passing algorithms fail in brain connectome analysis?}
\end{center}
To answer this question, we study the traditional brain connectomics pipeline (Figure~\ref{fig:pipeline}) as a statistical estimation problem.
The true connectome of each subject is latent, and we only observe an empirical connectivity matrix estimated from a finite recording, so the analysis is naturally framed as an errors-in-variables (EIV) problem. The empirical connectivity matrix, either the sample covariance $\hat{\Sigma}$ or the sample correlation $\hat{C}$, is used twice: as the node features and as the basis for the graph $\hat{A} = f(\hat{\Sigma})$ or $\hat{A} = f(\hat{C})$. 

\textbf{Main contributions.} Our contributions are summarized as follows:
\begin{itemize}[leftmargin=*, nosep, topsep=1pt]
    \item \textbf{An EIV formulation that allows untangling different effects.} Under an errors-in-variables formulation, we show in Section~\ref{sec:stats_framework} that the effect of graph convolution on the predictive performance of the method splits into three parts: \textbf{(i) a  graph smoothing effect} that induces a bias-variance tradeoff even with a correctly specified (oracle) graph; \textbf{ (ii) the graph estimation error}, which is the cost of smoothing over a wrongly estimated graph; and  \textbf{(iii) a double-dipping effect} that couples the graph operator with the feature noise when both come from the same sample. 
    \item \textbf{When convolution helps.} The balance of these three effects, and hence whether graph convolution improves on the raw features, depends on three ingredients: the \textbf{community structure} of the connectome, which describes how sharply ROIs group into communities; the \textbf{predictive signal}, which describes how the target aligns with these communities; and the \textbf{threshold}, which determines which connections become edges of the graph. We make all three explicit through a latent community model, building on the standard neuroscience assumption that ROIs belong to a small number of communities (Section~\ref{sec:latent_community}). Under this model, convolution improves on the raw features only when its noise reduction outweighs the signal it distorts, which restricts the gains to short recordings. Simulations, where the true connectome is known, validate the theory (Section~\ref{sec:simulations}).
    \item \textbf{Practical relevance for real data.} On fMRI data, we build proxies for the oracle and independent graphs to separate the three effects in practice, and we show how the theory can guide the choice of graph, threshold and window length (Section~\ref{sec:real_data}, Appendix~\ref{app:recommendations}).
\end{itemize}

\section{Related literature}\label{sec:rel_lit}
\paragraph{Graph neural networks for brain connectomes.} The brain connectome is commonly modeled as a network of interconnected regions~\citep{Bullmore2009} whose activity is driven by a small number of latent components~\citep{calhoun2001method,beckmann2004probabilistic,beckmann2005investigations}. Message passing with GCN, GraphSAGE or GAT convolutions~\citep{kipf2017semisupervisedclassificationgraphconvolutional,hamilton2018inductiverepresentationlearninglarge,velickovic2018graphattentionnetworks} underlies most connectome-specific models~\citep{li2021braingnn,febrinanto2023balanced,xu2025brainprompt,xu2025brainood} and the standard benchmarks~\citep{Cui_2023,said2024neurographbenchmarksgraphmachine}. In these pipelines the graph is obtained by thresholding the same empirical connectivity matrix that serves as the node features.

\paragraph{Evidence against message passing.} Graph-agnostic classifiers on the connectivity matrix match or outperform GNNs and graph transformers~\citep{yang2025do,han2025rethinkingfunctionalbrainconnectome}. On the NeuroGraph benchmarks, accuracy is unchanged when edges are randomly removed, even when all of them are removed~\citep{lararangel2025limitsapplyinggraphtransformers}. None of these studies explains the failure.

\paragraph{Theory of message passing.} Analyses of message passing~\citep{battaglia2018relationalinductivebiasesdeep} have focused on node-level tasks. Once linearized, convolutions act as low-pass filters~\citep{wu2019simplifying,nt2019revisiting}. This explains why models without nonlinearities retain most of their accuracy~\citep{wu2019simplifying,he2020lightgcn}, while repeated aggregation leads to oversmoothing~\citep{li2018deeperinsightsgraphconvolutional,rusch2023surveyoversmoothinggraphneural}. Closest to our setting, \citet{keriven2022littlemuchtheoreticalanalysis} shows on a latent-space random graph that a finite number of mean-aggregation steps improves node-level regression before oversmoothing sets in. Our setting differs in two ways: the task is graph-level, and the graph is estimated from the same noisy data as the features.

\paragraph{Measurement error and double dipping.} Predicting from a noisy estimate of the connectome is an errors-in-variables problem. Section~\ref{sec:stats_framework} reviews the corrections for covariate measurement error and explains why they do not apply here. In the standard pipeline the same window also selects the edges, which we call double dipping, and we quantify its effect for graph convolutions.

\section{Statistical framework}\label{sec:stats_framework}

 \begin{figure}
     \centering
     \includegraphics[width=\linewidth]{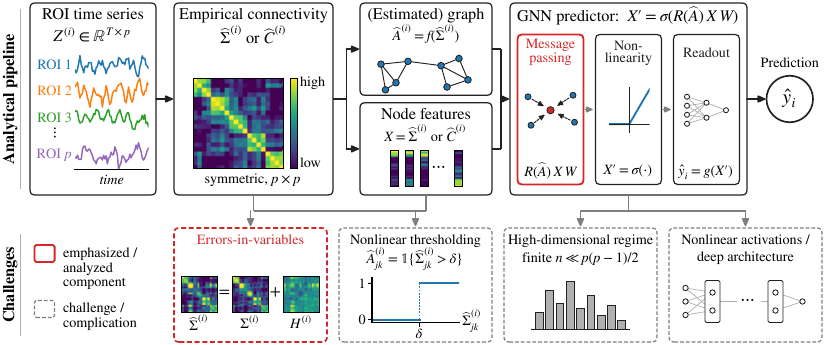}
     \caption{\textbf{Brain connectome GNN pipeline and statistical challenges.} \emph{Top:} the standard analytical pipeline, from ROI time series to prediction $\widehat{y}_i$. \emph{Bottom:} statistical challenges induced by the pipeline. The errors-in-variables component (red) is analyzed in this paper; the remaining challenges (gray, dashed) are left to future work.}
     \label{fig:pipeline}
 \end{figure}

For each subject $i = 1, \ldots, n$, let $Z^{(i)} \in \mathbb{R}^{T \times p}$ denote the centered fMRI time series measured at $p$ regions of interest (ROIs) over $T$ time points.
To keep the analysis simple, we assume independent time-points $Z^{(i)}_t \overset{\mathrm{i.i.d.}}{\sim} \mathcal N(0,\Sigma^{(i)})$, $t=1,\ldots,T$,  
where $\Sigma^{(i)}\in\mathbb{R}^{p\times p}$ is the subject's latent functional connectome. The connectome is not observed directly. Instead, we observe the empirical covariance
\begin{equation}
    \widehat\Sigma^{(i)} = \frac{1}{T}\sum_{t=1}^{T} Z^{(i)}_t Z^{(i)\top}_t
    = \Sigma^{(i)} + H^{(i)},
    \label{eq:sample-covariance}
\end{equation}
where $H^{(i)}$ is the finite-sample estimation error. For simplicity, we focus on the sample covariance $\widehat\Sigma^{(i)}$ throughout the theoretical analysis; extensions to the sample correlation matrix $\widehat C^{(i)}$, which is the more common choice in fMRI pipelines, are analyzed computationally in Appendix~\ref{app:simulations} and behave analogously. We study a simple supervised regression problem in which the patient-level target $y_i$ depends on the latent connectome through some (potentially non-linear) function $f^\star$:
\begin{equation}
    y_i = f^\star(\Sigma^{(i)}) + \varepsilon_i,
    \qquad
    \varepsilon_i\sim\mathcal N(0,\sigma_\varepsilon^2).
    \label{eq:true-linear-model}
\end{equation}
The purpose of brain connectomics analysis is thus to find the best approximation of $f^\star$ within some function class $\mathcal{F}$, typically parametrized by a vector of parameters $\theta$:
\begin{equation}
\hat{\theta} \in \text{argmin}_{f_{\theta} \in \mathcal{F}} \frac{1}{n} \sum_{i=1}^n \| y_i - f_{\theta}(\Sigma^{(i)})\|^2.
    \label{eq:true-linear-model2}
\end{equation}
\paragraph{An errors-in-variables regression problem.}Since the true connectivity $\Sigma^{(i)}$ is not observed, a simple solution is to use the estimated covariance matrix (also known as the plug-in estimator) $\widehat{\Sigma}^{(i)}$ and solve Eq.~\ref{eq:true-linear-model2}. However, predicting $y_i$ from $\hat{\Sigma}^{(i)}$ rather than $\Sigma^{(i)}$ places us in an \emph{errors-in-variables} regression setting. 
For simplicity, we restrict the function class to linear maps, $\mathcal{F} = \{f_\theta : x \mapsto x^\top \theta\}$, so that $y_i = \operatorname{vec}(\Sigma^{(i)})^\top \beta^\star + \varepsilon_i$. In this setting, the plug-in OLS estimator has a closed form. Write $x_i=\operatorname{vec}(\Sigma^{(i)})$ for the true features, $\eta_i=\operatorname{vec}(H^{(i)})$ for the noise, and $\hat x_i=x_i+\eta_i$ for the observed features. Under the Gaussian model above, the noise $\eta_i$ has conditional mean zero given $\Sigma^{(i)}$ and its conditional covariance scales as $T^{-1}$:
\begin{equation}
    \mathbb E_{\eta_i}[\eta_i\mid\Sigma^{(i)}]=0,\qquad
    \mathrm{Cov}_{\eta_i}(\eta_i\mid\Sigma^{(i)})=\frac{1}{T}\Psi^{(i)},\qquad
    \Psi^{(i)}=(I_{p^2}+K_{p,p})(\Sigma^{(i)}\otimes\Sigma^{(i)}),
    \label{eq:noise}
\end{equation}
where $K_{p,p}$ is the commutation matrix (Definition~\ref{def:commutation}). Let $S=\mathbb E_{\Sigma}[xx^{\top}]$ and $\Gamma=\mathbb E_{\Sigma}[\Psi]$, and write $\dagger$ for the Moore--Penrose pseudo-inverse. A classical errors-in-variables computation (Remark~\ref{thm:raw1} in Appendix~\ref{app:stats_framework}) gives the population OLS coefficient $\beta_{\mathrm{raw}}^*(T)=\Lambda(T)\beta^\star$, with $\Lambda(T):=(S+\tfrac{1}{T}\Gamma)^{\dagger}S$. Its test error is
\begin{equation}
    \mathrm{MSE}_{\mathrm{raw}}(T)
    =
    \beta^{\star\top}(I-\Lambda(T))^\top S(I-\Lambda(T))\beta^\star
    +
    \frac{1}{T}\beta^{\star\top}\Lambda(T)^\top\Gamma\Lambda(T)\beta^\star
    +\sigma_\varepsilon^2 .
    \label{eq:mse_raw}
\end{equation}
The first term is a bias: the noise pulls the coefficient away from $\beta^\star$. The second is tied to the variance: the noise in the features is carried into the prediction. Both terms depend on $T$, not on $n$, so it is important to note that more subjects do not remove the error; only longer recordings do. $\mathrm{MSE}_{\mathrm{raw}}$ is the baseline we compare every graph against in the rest of the paper. A convolution is appealing because averaging neighboring entries of $\widehat\Sigma$ can reduce the noise term, but it can also distort the signal. The next sections quantify this trade-off.

A large literature has studied corrections for covariate measurement error, spanning corrected high-dimensional Lasso objectives \citep{loh2011high,datta2017cocolasso}, score, likelihood, and deconvolution-based approaches for generalized and nonlinear models \citep{stefanski1987conditional,chen2011nonlinear,schennach2013measurement}, and recent nonparametric or neural approaches \citep{hu2022measurement}. None of these are directly applicable here: the measurement error in $\operatorname{vec}(\widehat\Sigma)$ is not exogenous with fixed covariance but dense, entry-correlated, and dependent on the unobserved latent connectome (Eq.~\ref{eq:noise}). 

\paragraph{The appeal of the graph.}

An alternative is to exploit the inherent problem structure: in neuroscience, each ROI is typically assumed to participate in $K \leq p$ latent neural communities \citep{Bullmore2009} that can be represented by a graph. For this reason, GNN-based algorithms have been extensively used in this field as standalone standard convolutions \citep{kipf2017semisupervisedclassificationgraphconvolutional, hamilton2018inductiverepresentationlearninglarge, velickovic2018graphattentionnetworks} or as part of brain-connectome-tailored architectures \cite{said2024neurographbenchmarksgraphmachine, li2021braingnn}. Their goal is twofold: (a) obtaining a richer feature map and (b) denoising the empirical covariance matrix by averaging over appropriate neighborhoods. We postulate that the main appeal of these methods in our setting stems from the denoising effect: the usually reported strengths of GNNs as feature maps (such as learning permutation-invariant representations) do not apply here, since ROIs are aligned across patients and therefore non-exchangeable. 

A graph message passing layer is defined as $X' = \sigma(R(A)\, X\, W)$, where $A \in \mathbb{R}^{p \times p}$ is an adjacency matrix, $X \in \mathbb{R}^{p \times m}$ a node feature matrix, $W \in \mathbb{R}^{m \times m'}$ learnable weights, $R(A)$ the graph operator induced by $A$ (e.g.\ $\widetilde D^{-1}\widetilde A$ for GraphSAGE~\cite{hamilton2018inductiverepresentationlearninglarge}, $\widetilde D^{-1/2}\widetilde A\,\widetilde D^{-1/2}$ for GCN \cite{kipf2017semisupervisedclassificationgraphconvolutional}, with $\tilde{A}=A + I$ and $\tilde{D}=\operatorname{diag}(\tilde{A} \mathbf{1})$), and $\sigma$ an activation function. 

The brain connectomics pipeline (Figure~\ref{fig:pipeline}) proceeds as follows. First, each subject's time series is split into windows of length $T$, so that shorter windows yield more samples per subject. Second, a connectivity matrix is estimated from each window (usually the correlation $\widehat C^{(i)}$ rather than the covariance $\widehat\Sigma^{(i)}$). Third, this matrix is used twice: as the node features, $X = \widehat\Sigma^{(i)}$ or $X = \widehat C^{(i)}$, and to build the graph $\widehat A$ through a non-linear function $f$ (for instance, hard or absolute-value thresholding \citep{said2024neurographbenchmarksgraphmachine}, either at a fixed level, $\widehat A_{jk}=\mathbbm 1\{\widehat\Sigma_{jk}>\delta\}$ with $\delta\in\mathbb R$, or at a fixed edge density, $\widehat A_{jk}=\mathbbm 1\{\widehat\Sigma_{jk}>q_{1-\tau}(\widehat\Sigma)\}$ with $\tau\in(0,1)$, where $q_{1-\tau}$ is the $(1-\tau)$-quantile of the off-diagonal entries). Finally, one or more message-passing layers, usually with non-linear activations and followed by a readout, are trained on the $n$ windows pooled across subjects, and $n$ is often smaller than the number $p(p-1)/2$ of pairwise interactions between the $p$ ROIs.

We focus here on understanding the denoising effect and the bias--variance tradeoff induced by the convolution, which requires disentangling it from the other factors summarized at the bottom of Figure~\ref{fig:pipeline}. For the purposes of this paper, we therefore simplify the pipeline. We use the covariance and independent time points, as introduced above. We work in an infinite-$n$ (asymptotic) regime, where $T$ is a free parameter rather than tied to $n$ through the windowing. We also restrict our analysis to linear models applied to the vectorized smoothed covariance, with the graph obtained by thresholding at a fixed level $\delta$ rather than a fixed density $\tau$. The linear restriction aligns with much of the empirical work that uses only the message-passing component and does not incorporate additional non-linearities \citep{wu2019simplifying,nt2019revisiting,he2020lightgcn}. Under these simplifications, the population risk has a closed form, and the analysis still captures how the message-passing layer interacts with the measurement noise. Appendix~\ref{app:progressive_relaxation} relaxes these simplifications one at a time on real data.

We compare prediction based on the raw feature map $\widehat{x} = \operatorname{vec}(\widehat{\Sigma}^{(i)})$ with that on the smoothed feature map $\widetilde{x}_{A} = \operatorname{vec}(R(A)\, \widehat{\Sigma}^{(i)})$, where $R(A)\widehat{\Sigma}^{(i)}$ is a linear message-passing convolution (e.g.\ $R(A) = \tilde{D}^{-1} \tilde{A}$ for GraphSAGE). We consider four choices of $A$:
\begin{enumerate}[leftmargin=1.5em, nosep,topsep=0pt]
    \item A \emph{population graph} $(A^{\mathrm{pop}})_{jk}(\delta) = \mathbbm 1\{(\Sigma_0)_{jk} > \delta\}$, where $\Sigma_0 = \mathbb{E}_\Sigma[\Sigma^{(i)}]$, is the average brain connectivity, independent of each subject's feature noise.
    \item The \emph{oracle graph} $A^{\text{oracle} (i)}_{jk}(\delta) = \mathbbm 1\{\Sigma^{(i)}_{jk} > \delta\}$, constructed from the true $\Sigma^{(i)}$, which captures the ``best-case'' impact of smoothing under correct graph specification.
    \item The \emph{same-sample graph} $\widehat A^{(i)}_{jk}(\delta) = \mathbbm 1\{\widehat\Sigma^{(i)}_{jk} > \delta\}$, built from the same empirical covariance as the node features; the most common choice in practice \citep{li2021braingnn,febrinanto2023balanced,xu2025brainprompt,xu2025brainood}.
    \item An \emph{independent graph} $ A^{\perp (i)}_{jk}(\delta) = \mathbbm 1\{\widehat\Sigma^{\perp(i)}_{jk} > \delta\}$, where $\widehat\Sigma^{\perp(i)} = \Sigma^{(i)} + H^{\perp(i)}$ with $H^{\perp(i)} \indep H^{(i)} \mid \Sigma^{(i)}$. It has graph-estimation error but the errors in the graph and features are independent.
\end{enumerate}

Replacing $\widehat x$ with the smoothed features $\widetilde x_A=\mathcal{R}_A\widehat x$, where $\mathcal{R}_A=I_p\otimes R(A)$ is the vectorized convolution, keeps us in an errors-in-variables regression, now with the convolution sitting between the noise and the coefficient. 
For any graph construction $A$, whether fixed or random, the population OLS test error is
\[
    \mathrm{MSE}_A(T)=\mathbb{E}_{\Sigma}[h(\Sigma)^2]-c_A^\top M_A^{\dagger}c_A+\sigma_\varepsilon^2,
\]
where $h(\Sigma)=\operatorname{vec}(\Sigma)^\top\beta^\star$ is the noiseless target, $M_A=\mathbb{E}_{\Sigma,\eta,A}[\widetilde x_A\widetilde x_A^\top]$ and $c_A=\mathbb{E}_{\Sigma,\eta,A}[\widetilde x_A\,h(\Sigma)]$. Writing $\widehat x=x+\eta$, both split into a signal part, a noise part, and cross terms between the two:
\[
\begin{aligned}
    M_A &= \mathbb{E}_{\Sigma,\eta,A}[\mathcal{R}_A xx^\top\mathcal{R}_A^\top]
    + \mathbb{E}_{\Sigma,\eta,A}[\mathcal{R}_A x\eta^\top\mathcal{R}_A^\top]
    + \mathbb{E}_{\Sigma,\eta,A}[\mathcal{R}_A \eta x^\top\mathcal{R}_A^\top]
    + \mathbb{E}_{\Sigma,\eta,A}[\mathcal{R}_A \eta\eta^\top\mathcal{R}_A^\top], \\
    c_A &= \mathbb{E}_{\Sigma,\eta,A}[\mathcal{R}_A xx^\top]\beta^\star + \mathbb{E}_{\Sigma,\eta,A}[\mathcal{R}_A \eta x^\top]\beta^\star.
\end{aligned}
\]
How these terms behave depends on how the graph relates to the feature noise, which gives three cases (Proposition~\ref{prop:eiv-message-passing}, Appendix~\ref{app:theory_eiv}):
\begin{description}[nosep, topsep=0em, leftmargin=0.5em]
    \item[(i) For the same-sample graph $\widehat A$,] the graph is built from the same $\widehat\Sigma$ as the features, so it depends on the noise it is meant to average out. All terms are present, and the cross terms do not vanish due to this double-dipping effect. 
    \item[(ii) For an independent graph that varies across subjects,]  such as $A^{\text{oracle}}$and $A^{\perp}$, the graph is independent of the feature noise given $\Sigma$, so the cross terms vanish.
    \item[(iii) For a fixed graph shared across samples,] as for $A^{\text{pop}}$, the same convolution is applied to every subject, so $\mathcal{R}_A$ is a fixed linear map. A linear model on $\widetilde x_A$ is then still a linear model on $\widehat x$, and message passing can never improve on the raw features. It only ties with the raw features when $R(A)$ is invertible and can only do worse otherwise (Proposition~\ref{prop:no_linear_gain}, Appendix~\ref{app:theory_eiv}).

\end{description}

In the two cases where the graph varies across subjects, the sign of each effect is not determined in general. It depends on how the connectome $\Sigma$ is structured, on how the target relates to that structure, and on how the graph is built. In the next section we introduce a latent community model that makes these three ingredients explicit.

\section{Brain connectome inspired latent communities model}
\label{sec:specialization}
\label{sec:latent_community}
We now specialize the previous results to a simple connectome
model, showing how the favorable regime changes as we move
from the oracle graph $A^{\text{oracle}}$, to the independent graph
$A^{\perp}$, and to the same-sample graph $\widehat A$,
with the goal of isolating respectively the
\textbf{graph smoothing effect}, the
\textbf{graph estimation error}, and the
\textbf{double-dipping effect}.

The model represents brain activity through a small number of
interacting latent components~\citep{calhoun2001method,
beckmann2004probabilistic,beckmann2005investigations,airoldi2007mixedmembershipstochasticblockmodels}.
Let $F^{(i)}\in\mathbb R^{p\times d}$ collect the ROIs' loadings
onto $d$ latent dimensions. Each ROI $j$ belongs to one community
$z_j\in\{1,\ldots,K\}$, $K\le\min\{p,d\}$; the communities are balanced, with
$|C_k|=q=p/K$ and $C(j):=C_{z_j}$.
Following Appendix~\ref{app:lcm_structure}, we take fixed,
orthogonal centroids $\mu_k$ with $\|\mu_k\|^2=\phi^2d$.
For each subject, let $f_j^{(i)}=\mu_{z_j}+\nu_j^{(i)}$, where $\nu_j^{(i)}\overset{\mathrm{i.i.d.}}{\sim}\mathcal N(0,\sigma^2I_d)$ across ROIs $j$ and subjects $i$, and define $\Sigma^{(i)}=F^{(i)}F^{(i)\top}+\psi^2I_p$, with $f_j^{(i)\top}$ the $j$-th row of $F^{(i)}$ and
the perturbations are independent across ROIs and subjects.
The ROI partition, centroids, latent dimension $d$, and parameters $\phi,\psi>0$ are fixed across subjects.
Under this model, we characterize three ingredients.

\begin{figure}
    \centering
    \includegraphics[width=\linewidth]{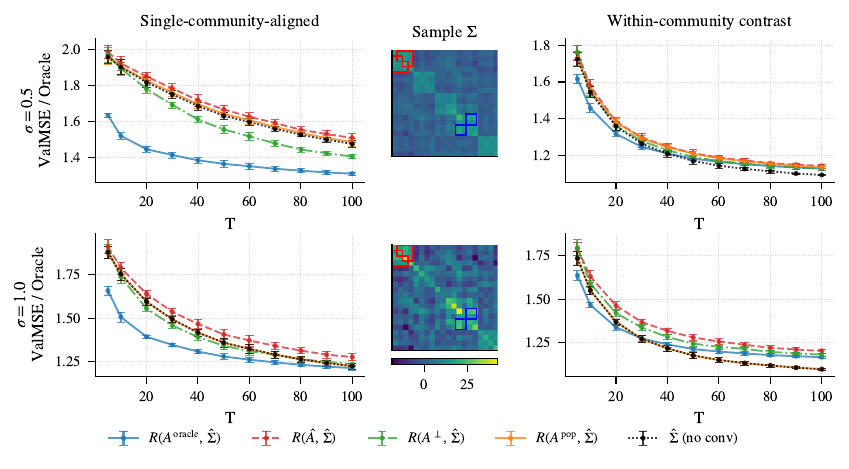}
    \caption{\textbf{Decoupling the effects for different values of $\beta$ and latent structures.} Center: true covariance $\Sigma$ for two graph regimes ($\sigma$); overlays mark the support of two coefficient patterns ($\beta$). Left/right: normalized ValMSE vs.\ $T$ for $\mathrm{SageMean}(A, \widehat\Sigma)$ and a no-conv baseline. Mean $\pm$ 1 SD over 5 seeds. $p=20$, $K=5$, $n=25{,}000$; thresholds $\delta_{\sigma=0.5}\approx 9.10$, $\delta_{\sigma=1.0}\approx 10.20$.}
    \label{fig:main}
\end{figure}

\textbf{Community structure, $\sigma$.}
The parameter $\sigma$ controls subject variability around the
shared community pattern. For $j\ne k$, the within- and
between-community covariance means and common variance are $\mu_W=\phi^2d$, $\mu_B=0$, and $s^2(\sigma)=2\sigma^2\phi^2d+\sigma^4d$, respectively
(Lemma~\ref{lem:lcm_moments}).
Increasing $\sigma$ increases the spread around these fixed
levels as shown in the heatmaps in Figure \ref{fig:main}. 

\textbf{Predictive signal, $B^\star$.}
The symmetric coefficient $B^\star\in\mathbb R^{p\times p}$
specifies $h(\Sigma)=\langle B^\star,\Sigma\rangle$, with
$\beta^\star=\operatorname{vec}(B^\star)$.
We characterize its alignment with the same communities through $
    \max_j\max_{\ell\in C(j)}
    \sum_{k=1}^p(B^\star_{jk}-B^\star_{\ell k})^2
    \le\kappa_\beta^2$.
Small $\kappa_\beta$ describes community-smooth predictive
profiles; the ideal case $\kappa_\beta=0$ gives coefficients
depending only on $(z_j,z_k)$.
Community-level, single-edge, and contrast targets are defined
in Appendix~\ref{app:lcm_alignment}.

\textbf{Threshold, $\delta$.}
The threshold selects averaging neighborhoods from these
connectomes. For the oracle,
$ A_{jk}^{\text{oracle}(i)}(\delta)=
\mathbbm1\{\Sigma_{jk}^{(i)}>\delta\}$ for $j\ne k$,
with mean aggregation $R(A)=\widetilde D^{-1}(A+I_p)$.
At the boundaries, $\delta=+\infty$ gives the empty graph and $R=I_p$, the raw features, while $\delta=-\infty$ gives the complete graph and $R=\mathbf1\mathbf1^\top/p$, the global average.
The interesting case is an intermediate threshold, $\mu_B<\delta<\mu_W$, at distance $\gamma_\delta:=\min\{\mu_W-\delta,\delta-\mu_B\}$ from the two means. The graph then mostly links ROIs of the same community, and the operator splits as
\[
    R(A^{\text{oracle}})=P_C+\Delta R_\delta,
    \qquad
    (P_C)_{j\ell}=q^{-1}\mathbbm1\{\ell\in C(j)\}.
\]
Here $P_C$ averages each ROI over its community, and $\Delta R_\delta$ collects the subject-specific deviations, which depend on $\sigma$ and $\delta$.

The following results examine how these three ingredients
jointly affect prediction as the measurement length $T$ varies.

\begin{table}
\caption{Graph constructions and prediction conclusions. Comparisons concern unregularized population linear prediction under the stated assumptions.}
\label{tab:graph_hierarchy_latent}
\centering
\small
\begin{tabular}{p{0.18\linewidth}p{0.25\linewidth}p{0.48\linewidth}}
\toprule
\textbf{Graph} & \textbf{Effects isolated} & \textbf{Main prediction conclusion} \\
\midrule
$A^{\mathrm{pop}}$ (population) & Fixed smoothing &
Cannot improve optimized raw prediction. \\[0.3em]
$A^{\text{oracle}}$ (oracle) & Subject-specific smoothing &
Can help when variance reduction outweighs squared bias and raw refitting. \\[0.3em]
$A^{\perp}$ (independent) & $+$ Graph estimation &
Same tradeoff, with an additional graph-error bound. \\[0.3em]
$\widehat A$ (same-sample) & $+$ Noise reuse &
Close to independent prediction when graph estimation is reliable and regression is stable. \\
\bottomrule
\end{tabular}
\end{table}

\result{Oracle graph smoothing effect}
\textit{ Smoothing over the oracle graph reduces measurement noise but also distorts the signal, and the balance between the two depends on the predictive signal $B^\star$, the community structure $\sigma$ and the threshold $\delta$. If smoothing introduces prediction error that persists even after refitting the linear predictor on noiseless features, it cannot outperform raw features for sufficiently long recordings.}

The following proposition is a more formal version of Result 1. 
\begin{proposition}[Oracle smoothing]
\label{prop:oracle_latent_community}
Under the EIV assumptions of Remark~\ref{thm:raw1} (Appendix~\ref{app:stats_framework}), fix any threshold $\delta$; the oracle operator $R\left(A^{\text{oracle}}(\delta)\right)$ is a function of $\Sigma$.
Assume all displayed moments are finite and define the squared bias and the variance reduction:
\begin{align*}
    \mathcal B_{A^{\text{oracle}}}(\delta)
    &:=
    \mathbb E_\Sigma
    \left[\langle B^\star,R\left(A^{\text{oracle}}(\delta)\right)\Sigma-\Sigma\rangle^2\right],\\
    \Delta\mathcal V_{A^{\text{oracle}}}(\delta)
    &:=
    T\,\mathbb E_\Sigma\left[
        \operatorname{Var}_\eta
        \{\langle B^\star,H\rangle\mid\Sigma\}
        -
        \operatorname{Var}_\eta
        \{\langle B^\star,R\left(A^{\text{oracle}}(\delta)\right) H\rangle\mid\Sigma\}
    \right].
\end{align*}

\textbf{Sufficient condition.}
The population MSEs satisfy, for every $T>0$,
\[
    \operatorname{MSE}_{\mathrm{oracle}}(T,\delta)
    -\operatorname{MSE}_{\mathrm{raw}}(T)
    \le
    \mathcal B_{A^{\text{oracle}}}(\delta)
    -\frac{\Delta\mathcal V_{A^{\text{oracle}}}(\delta)}{T}
    +r_{\mathrm{raw}}(T),
\]
where
\[
    r_{\mathrm{raw}}(T):=
    T^{-2}\beta^{\star\top}
    \Gamma(S+T^{-1}\Gamma)^\dagger\Gamma\beta^\star
    \ge0.
\]
If the support of $\Sigma$ spans the symmetric matrices (Remark~\ref{thm:raw1}), as in the latent community model with $\sigma>0$, then $r_{\mathrm{raw}}(T)=O(T^{-2})$.
Thus oracle smoothing improves whenever
$\mathcal B_{A^{\text{oracle}}}(\delta)+r_{\mathrm{raw}}(T)
<\Delta\mathcal V_{A^{\text{oracle}}}(\delta)/T$.

\textbf{Necessary condition.}
Let $b_\delta:=
    \min_{\gamma\in\mathbb R^{p^2}}
    \mathbb E_\Sigma\left[
        \left\{
        h(\Sigma)-\gamma^\top\operatorname{vec}(R\left(A^{\text{oracle}}(\delta)\right)\Sigma)
        \right\}^2
    \right]$
be the minimum noiseless prediction error, with $\gamma$ shared
across subjects. If $b_\delta>0$, then
\[
    \operatorname{MSE}_{\mathrm{oracle}}(T,\delta)
    <\operatorname{MSE}_{\mathrm{raw}}(T)
    \quad\Longrightarrow\quad
    T<\frac{\beta^{\star\top}\Gamma\beta^\star}{b_\delta}.
\]
\end{proposition}
The sufficient condition guarantees improvement when the variance reduction
exceeds the squared bias plus $r_{\mathrm{raw}}(T)$,
which accounts for refitting the raw regression.
If a positive error $b_\delta$ remains even after fitting on
noiseless smoothed features, sufficiently long recordings make
raw prediction more accurate, giving the necessary upper limit on $T$.

For the intermediate threshold, $\mu_B<\delta<\mu_W$, the decomposition introduced above gives
\[
    \mathcal B_{A^{\text{oracle}}}(\delta)
    =
    \mathbb E_\Sigma\!\left[
        \left\{
        \langle B^\star,(P_C-I_p)\Sigma\rangle
        +
        \langle B^\star,\Delta R_\delta\Sigma\rangle
        \right\}^{2}
    \right].
\]
Exact community smoothness, $P_CB^\star=B^\star$, removes
the first term, but not necessarily the second.
Thus $B^\star$ determines which information averaging must
preserve, while $\sigma$ and $\delta$ jointly determine the
subject-specific neighborhoods and their noise propagation.
We also note that strong concentration relative to the threshold margins makes
the oracle nearly fixed, unable to improve
optimized prediction (Proposition~\ref{prop:no_linear_gain}).
Clear communities and an aligned target therefore do not
alone guarantee a gain. Concrete examples for community-strength and single-edge targets
are given in Appendix~\ref{app:case-oracle}.

We now ask whether a graph estimated from an independent sample of the same subject keeps the gain of the oracle.

\result{Graph estimation error}\label{res:2} \textit{Replacing the oracle with a graph estimated from an independent sample adds a graph-estimation error bound $\mathcal E(T,\delta)$ that vanishes with the probability of an edge error. The independent graph still improves on the raw features when this bound is smaller than the gain of the oracle. Recovering the community partition shared by all subjects is not enough, because the gain comes from the subject-specific neighborhoods.}

The independent graph thresholds a second estimate of the same connectome, $A^{\perp}_{jk}(\delta) = \mathbbm 1\{\widehat\Sigma^\perp_{jk}>\delta\}$ with $\widehat\Sigma^\perp=\Sigma+H^\perp$ and $H^\perp\indep H\mid \Sigma$. Its edges can be wrong, which adds graph-estimation error, but its noise is independent of the feature noise, so there is no double dipping.

\begin{proposition}[Graph-estimation error for the independent graph]
\label{prop:independent_graph_window}

Under the assumptions of Proposition~\ref{prop:oracle_latent_community}, let $A^\perp$ be built from a conditionally independent copy $H^\perp$ of $H$ given $\Sigma$, and assume $\mathbb E_\Sigma\|\Sigma\|_F^4<\infty$.
Define the edge-error probability and the graph-estimation error:
\[
    \rho(T,\delta):=\mathbb P(A^\perp\ne A^{\text{oracle}}),
    \qquad
    \mathcal E(T,\delta):=C(1+T^{-1})\sqrt{\rho(T,\delta)},
\]
where $C$ depends only on $p$, $\|B^\star\|_F$ and $\mathbb E_\Sigma\|\Sigma\|_F^4$, not on $T$ or $\delta$. Then, for every $T>0$,
\[
    \operatorname{MSE}_{\mathrm{ind}}(T,\delta)
    -\operatorname{MSE}_{\mathrm{raw}}(T)
    \le
    \mathcal B_{A^{\text{oracle}}}(\delta)
    -\frac{\Delta\mathcal V_{A^{\text{oracle}}}(\delta)}{T}
    +\mathcal E(T,\delta)+r_{\mathrm{raw}}(T).
\]
Thus the independent graph improves whenever
$\mathcal B_{A^{\text{oracle}}}(\delta)+\mathcal E(T,\delta)+r_{\mathrm{raw}}(T)
<\Delta\mathcal V_{A^{\text{oracle}}}(\delta)/T$.
\end{proposition}

For $\mu_B<\delta<\mu_W$, graph estimation must recover the subject-specific neighborhoods $P_C+\Delta R_\delta$. The target $B^\star$ determines how much a wrong edge costs, while $\delta$ and $\sigma$ determine how often one occurs. An edge is wrong when the noise in $\widehat\Sigma_{jk}$, of size $\sqrt{v/T}$, exceeds the margin $\gamma_\delta$ between $\delta$ and the nearest community mean, where $v$ bounds the scaled noise variance $T\operatorname{Var}_\eta(H_{jk}\mid\Sigma)=\Sigma_{jj}\Sigma_{kk}+\Sigma_{jk}^2$. Hence, when communities are well separated, $\rho$ decays exponentially in $T/T_{\mathrm{graph}}(\delta)$, with $T_{\mathrm{graph}}(\delta)=v/\gamma_\delta^2$ (Appendix~\ref{app:case-independent-copy}). Graph estimation is therefore reliable only once $T\gg T_{\mathrm{graph}}(\delta)$. Edge errors of this kind are visible in data: thresholded connectivity graphs built from short recordings contain spurious between-network edges. Longer recordings improve edge estimation but also reduce the raw-feature noise that smoothing aims to remove, so the gain is confined to an intermediate range of $T$.

\result{Double-dipping effect} \textit{Using the same sample for graph and features adds a signed coupling term $\mathcal D_{\rm dd}(T,\delta)$ to the constraints of Result~\ref{res:2}. When measurement noise rarely changes the oracle graph and the fitted regression coefficients remain bounded, the same-sample and independent graphs have similar prediction errors. The sign of $\mathcal D_{\rm dd}$ depends on the target.}

The same-sample graph $\widehat A_{jk}(\delta) = \mathbbm 1\{\widehat\Sigma_{jk}>\delta\}$ uses the same noise $H$ for edge selection and feature construction. We treat double dipping as a signed contribution because reusing the sample can either improve or worsen prediction compared with using an independent graph.

\begin{proposition}[Double-dipping contribution for the same-sample graph]
\label{prop:double_dipping_window}
Under the assumptions of Proposition~\ref{prop:independent_graph_window}, define the double-dipping contribution $\mathcal D_{\rm dd}(T,\delta):=\operatorname{MSE}_{\mathrm{same}}(T,\delta)-\operatorname{MSE}_{\mathrm{ind}}(T,\delta)$. Then, for every $T>0$,
\[
    \operatorname{MSE}_{\mathrm{same}}(T,\delta)
    -\operatorname{MSE}_{\mathrm{raw}}(T)
    \le
    \mathcal B_{A^{\text{oracle}}}(\delta)
    -\frac{\Delta\mathcal V_{A^{\text{oracle}}}(\delta)}{T}
    +\mathcal E(T,\delta)+\mathcal D_{\rm dd}(T,\delta)+r_{\mathrm{raw}}(T).
\]
Thus the same-sample graph improves whenever
$\mathcal B_{A^{\text{oracle}}}(\delta)+\mathcal E(T,\delta)+\mathcal D_{\rm dd}(T,\delta)+r_{\mathrm{raw}}(T)
<\Delta\mathcal V_{A^{\text{oracle}}}(\delta)/T$.
The contribution $\mathcal D_{\rm dd}$ is signed: negative values favor the same-sample graph.
\end{proposition}
The two graphs have the same conditional law given $\Sigma$ (Lemma~\ref{lem:marginal}), so $\mathcal D_{\rm dd}$ isolates noise reuse, not graph quality. For intermediate thresholds both estimate $P_C+\Delta R_\delta$, and genuine subject-specific variation is not double dipping. Under additional moment and spectral assumptions (Appendix~\ref{app:case-estimated-graph}), $|\mathcal D_{\rm dd}|\le C_{\rm dd}\sqrt{\rho/T}$. Double dipping therefore vanishes at $\delta=\pm\infty$ and is small whenever edge errors are rare, for instance once $T\gg T_{\mathrm{graph}}(\delta)$ with well-separated communities. In the intermediate regime an independent graph and same-sample graph have similar behavior.

\section{Simulations}
\label{sec:simulations}
We validate the theoretical predictions of Section~\ref{sec:specialization} under controlled conditions where the ground-truth covariance $\Sigma^{(i)}$ is known. We generate synthetic data from the latent community model (Appendix~\ref{app:latent}) to study how the five feature maps from Section~\ref{sec:stats_framework} behave as we vary the community structure, the predictive signal, and the threshold. As in the theoretical proofs, we use balanced communities with fixed orthogonal centroids and fixed latent dimension, so that the main latent-structure knob is the subject-level variability $\sigma$: small $\sigma$ gives strongly clustered connectomes, while large $\sigma$ weakens the community structure. For each subject, we sample a time series of length $T$, compute $\widehat\Sigma^{(i)}$ (deferring results for $\widehat C^{(i)}$ to Appendix~\ref{app:simulations}), and evaluate the various feature maps at threshold level $\delta$: raw, oracle graph, independent graph, same-sample graph, and population graph. Each is fit by OLS with an intercept on standardized features (80/20 train/test split). The dependent variable is generated using regression coefficients that span a range of favorability to graph smoothing.

\begin{figure}
    \centering
    \includegraphics[width=\linewidth]{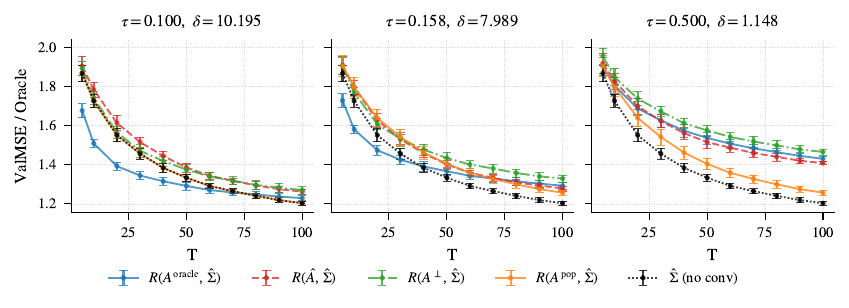}
    \caption{\textbf{Decoupling the effect of the threshold $\delta$.} Each panel fixes a different threshold $\delta$ on $\Sigma$ and plots normalized ValMSE vs.\ $T$ for $\mathrm{SageMean}(A, \widehat\Sigma)$ and the no-conv baseline. Mean $\pm$ 1 SD over 5 seeds. $\beta=\texttt{community\_aligned}$, $\sigma=1.0$, $p=20$, $K=5$, $n=25{,}000$.}
    \label{fig:tau-effect}
\end{figure}

\paragraph{Results.}\label{subsec:sim_results}
We disentangle the three effects of Section~\ref{sec:specialization} as gaps between the feature maps $\{\widehat\Sigma,\,R(A^{\text{oracle}})\widehat\Sigma,\,R(A^{\perp})\widehat\Sigma,\,R(\widehat A)\widehat\Sigma,\,R(A^{\mathrm{pop}})\widehat\Sigma\}$, normalized by the oracle MSE.

\providecommand{\pmse}[1]{{\scriptsize$\pm$#1}}
\providecommand{\msd}[3]{$#1\rlap{$^{#2}_{\pm#3}$}$\hphantom{$^{***}_{\pm.000}$}}
\providecommand{\prov}[1]{\textcolor{red}{#1}}
\providecommand{\gs}[2]{$#1\rlap{$^{#2}$}$\hphantom{$^{***}$}}

\begin{table}
\caption{\textbf{Decoupling the effects on real data.} Subject-level test accuracy for autism diagnosis on ABIDE (Harvard-Oxford, $p=106$) with $\mathrm{SageMean}(A,\widehat C)$, one convolution, identity activation,  $n$ fixed across $T$, $\tau=0.10$. Mean $\pm$ 1 SD over 20 seeds; paired $t$-test over seeds against $\widehat C$ (left) or zero (right): $^{*}p<0.10$, $^{**}p<0.05$, $^{***}p<0.01$. The majority-class predictor has accuracy 0.50.}
\label{tab:abide_scenarios}
\centering
\footnotesize
\setlength{\tabcolsep}{0pt}
\begin{tabular*}{\linewidth}{@{\extracolsep{\fill}} c ccccc cccc}
\toprule
$T$ & $\widehat C$ & $A^{\text{oracle}}_{\rm est}$ & $A^{\perp}_{\rm est}$ & $\widehat A$ & $A^{\mathrm{pop}}$ & $A^{\text{oracle}}_{\rm est}-\widehat C$ & $A^{\perp}_{\rm est}-A^{\text{oracle}}_{\rm est}$ & $\widehat A-A^{\perp}_{\rm est}$ & $A^{\mathrm{pop}}-\widehat C$ \\
\midrule
5 & \msd{0.560}{}{.033} & \msd{0.582}{***}{.034} & \msd{0.561}{}{.031} & \msd{0.563}{}{.035} & \msd{0.562}{}{.039} & \gs{+0.022}{***} & \gs{-0.021}{***} & \gs{+0.002}{} & \gs{+0.001}{} \\
15 & \msd{0.625}{}{.040} & \msd{0.630}{}{.033} & \msd{0.619}{}{.040} & \msd{0.619}{}{.039} & \msd{0.618}{}{.032} & \gs{+0.004}{} & \gs{-0.010}{*} & \gs{+0.000}{} & \gs{-0.007}{} \\
25 & \msd{0.651}{}{.037} & \msd{0.668}{**}{.030} & \msd{0.641}{}{.047} & \msd{0.650}{}{.044} & \msd{0.653}{}{.037} & \gs{+0.017}{**} & \gs{-0.027}{***} & \gs{+0.009}{*} & \gs{+0.002}{} \\
40 & \msd{0.682}{}{.034} & \msd{0.678}{}{.031} & \msd{0.669}{**}{.030} & \msd{0.663}{**}{.036} & \msd{0.672}{*}{.029} & \gs{-0.004}{} & \gs{-0.009}{**} & \gs{-0.006}{} & \gs{-0.010}{*} \\
\bottomrule
\end{tabular*}
\end{table}

We start by isolating the graph smoothing effect. The population graph $A^{\mathrm{pop}}$ tracks $\widehat\Sigma$ closely in Figure~\ref{fig:main}, consistent with Proposition~\ref{prop:no_linear_gain}, so the relevant comparison is $A^{\text{oracle}}$ vs.\ raw.
Proposition~\ref{prop:oracle_latent_community} predicts a bias-variance tradeoff whose benefit is confined to small $T$ whenever the noiseless smoothed features leave residual error, and Figure~\ref{fig:main} confirms this: for the community-smooth target (red overlay), $\mathcal B_{A^{\text{oracle}}}(\delta)$ is small and $R(A^{\text{oracle}})\widehat\Sigma$ improves prediction across all $T$; for the within-community contrast (blue overlay), smoothing averages together the ROIs whose difference is predictive, so the oracle helps only at small $T$.
Increasing $\sigma$ from $0.5$ to $1.0$ spreads the entries around the community levels and narrows the oracle gain. The threshold $\delta$ modulates this gain: at sparse, high-margin thresholds the oracle preserves the community structure and the smoothing benefit is maximal, while as $\delta$ moves toward the bulk the graph densifies and the oracle gain vanishes (Figure~\ref{fig:tau-effect}, right).

We next turn to the graph-estimation error, comparing $A^{\perp}$ to $A^{\text{oracle}}$. When the oracle does not help, $A^{\perp}$ tracks $A^{\text{oracle}}$ trivially. When it does (top-left panel, Figure~\ref{fig:main}), a large fraction of the oracle gain is lost to mis-estimated edges, matching Proposition~\ref{prop:independent_graph_window}. Figure~\ref{fig:tau-effect} traces this dependence in $\delta$: at a sparse, high-margin threshold (left), $A^{\text{oracle}}$ delivers a large gain and $A^{\perp}$ recovers part of it; as $\delta$ moves toward the bulk (right), the graph approaches complete and the oracle ceases to help (Proposition~\ref{prop:no_linear_gain}).

Finally, the double-dipping effect is captured by the gap between $\widehat A$ and $A^{\perp}$. Across Figure~\ref{fig:tau-effect}, $R(\widehat A)\widehat\Sigma$ tracks $R(A^{\perp})\widehat\Sigma$ closely, consistent with the high-margin regime of Proposition~\ref{prop:double_dipping_window} where few entries of $\Sigma$ lie within one Wishart noise scale of $\delta$. Pipelines with thresholds closer to the bulk, or weaker community separation, may incur a larger signed coupling penalty.

Extensions and additional results, including application to correlation matrices as well as tests of robustness under departures from the i.i.d.\ sampling assumption and at finite sample sizes, are presented in Appendix~\ref{app:simulations}.

\section{Real-data controlled experiments} \label{sec:real_data}
We now show that the theoretical findings, already validated in the simulations, extend to real data. The traditional brain connectomics pipeline \citep{said2024neurographbenchmarksgraphmachine, Cui_2023} is employed. From each subject's ROI time series we extract rolling windows of length $T$ and compute the per-window sample correlation matrix $\widehat C^{(i)}$, which serves both as the node feature matrix and as the basis for graph construction; the correlation rather than the covariance matrix is used to stay close to practice. Thresholding it yields the four graphs of Section~\ref{sec:stats_framework}: $\widehat A$ from the same window (the standard pipeline), $A^{\perp}_{\rm est}$ from an independent window of the same subject, $A^{\text{oracle}}_{\rm est}$ from the full-length series (a low-noise proxy for the unobserved oracle), and the population graph $A^{\text{pop}}$. Each graph is paired with the same minimal architecture: a single convolution followed by an activation (identity or ReLU) and a readout (mean or vectorization).

Real data depart from the assumptions of the theory and the simulations. Table~\ref{tab:abide_scenarios} retains three of them, which Appendix~\ref{app:real_data} relaxes one at a time. First, the theory treats $R(A)$ as linear, whereas practice applies a non-linearity; we compare identity and ReLU activations. Second, $n$ is finite and, with rolling windows, grows as $T$ shrinks; we hold $n$ fixed across $T$, isolating the window length, or let it vary as in practice. Third, the theory regresses on vectorized features, whereas practice pools node embeddings; we compare both readouts. In every experiment the oracle and independent graphs are estimated and training is regularized.

Table~\ref{tab:abide_scenarios} reproduces the simulation ordering on real data. The proxy oracle $A^{\text{oracle}}_{\rm est}$ improves on the raw features at $T=5$ and $T=25$ but not at $T=40$, so the favorable window closes at finite $T$ (Proposition~\ref{prop:oracle_latent_community}); the population graph $A^{\text{pop}}$, shared by all subjects, never helps. Graph-estimation error is significant at every $T$ and cancels the oracle gain (Proposition~\ref{prop:independent_graph_window}), whereas the double-dipping term is small, of no consistent sign and comparable to the estimation loss only at $T=40$ (Proposition~\ref{prop:double_dipping_window}), so the same-sample graph never beats the raw features.

Table~\ref{tab:abide_relaxation} in Appendix~\ref{app:real_data} relaxes these three assumptions cumulatively, and Tables~\ref{tab:abide_gat} to~\ref{tab:abide_tau} repeat the analysis with GAT, two other thresholds and two more datasets. The ReLU weakens the oracle gain without removing it, while freeing $n$ removes it from $T=15$ on and strengthens it at $T=5$, where short windows yield the most samples. Mean pooling then removes the gain left at $T=5$, so in the deployed pipeline no graph improves significantly on the raw features at any $T$, as in Figure~\ref{fig:abide_adhd_perf}.

\section{Conclusions}\label{sec:conclusion}

\paragraph{Contributions.}
We framed message passing on brain connectomes as an errors-in-variables regression and split its effect on prediction into three parts: graph smoothing, graph estimation error and double dipping. Under a latent community model, we showed how each part depends on the community structure, the predictive signal and the threshold. Even with the oracle graph, convolution helps only when its noise reduction outweighs the signal it distorts, which confines the gain to short recordings, and a graph shared by all subjects can never help. Simulations with the oracle, independent and same-sample graphs isolate each effect and confirm these predictions. On real data, proxies for the oracle and independent graphs reproduce the same picture: the proxy oracle improves on the raw correlation matrix at short windows and the gain disappears as $T$ grows, the population graph never helps, most of the oracle gain is lost to graph-estimation error, and double dipping stays small. This answers the question we started from: the standard pipeline fails to beat graph-agnostic methods because the graph it can estimate from a short window is too noisy to deliver the smoothing gain that the true graph would offer.

\paragraph{Limitations: from simulations to real data.}
Our theory is stated for a linear model in the population scenario ($n \to \infty$), the most favorable case for graph convolutions. Like the simulations, it relies on strict assumptions chosen to keep the mathematics tractable. ReLU and attention fall outside the linear theory but leave the real-data picture unchanged; a theory for them, for finite samples and for richer connectomes is left to future work. For the same reason, we do not provide an explicit rule for computing the minimum threshold on $T$. Finally, alternative methods for estimating the graph, beyond simple thresholding, are a relevant direction for future work.

\section*{Acknowledgments}

Part of this work was done while T.C. and J.L. were at the Department of Statistics, University of Chicago. C.D. would like to acknowledge support for this project
from the National Science Foundation (NSF grant IIS-2238616). J.L. would like to thank the support for this project from the National Science Foundation (NSF grant DMS-2515926). This work was completed in part with resources provided by the University of Chicago’s Research Computing Center.

\newpage

\bibliography{references}

\begin{thebibliography}{43}
\providecommand{\natexlab}[1]{#1}
\providecommand{\url}[1]{\texttt{#1}}
\expandafter\ifx\csname urlstyle\endcsname\relax
  \providecommand{\doi}[1]{doi: #1}\else
  \providecommand{\doi}{doi: \begingroup \urlstyle{rm}\Url}\fi

\bibitem[Bullmore and Sporns(2009)]{Bullmore2009}
Ed~Bullmore and Olaf Sporns.
\newblock Complex brain networks: graph theoretical analysis of structural and
  functional systems.
\newblock \emph{Nature Reviews Neuroscience}, 10\penalty0 (3):\penalty0
  186--198, 2009.
\newblock \doi{10.1038/nrn2575}.

\bibitem[Cui et~al.(2023)Cui, Dai, Zhu, Kan, Gu, Lukemire, Zhan, He, Guo, and
  Yang]{Cui_2023}
Hejie Cui, Wei Dai, Yanqiao Zhu, Xuan Kan, Antonio Aodong~Chen Gu, Joshua
  Lukemire, Liang Zhan, Lifang He, Ying Guo, and Carl Yang.
\newblock {BrainGB}: A benchmark for brain network analysis with graph neural
  networks.
\newblock \emph{IEEE Transactions on Medical Imaging}, 42\penalty0
  (2):\penalty0 493–506, February 2023.
\newblock ISSN 1558-254X.
\newblock \doi{10.1109/tmi.2022.3218745}.
\newblock URL \url{http://dx.doi.org/10.1109/TMI.2022.3218745}.

\bibitem[Said et~al.(2023)Said, Bayrak, Derr, Shabbir, Moyer, Chang, and
  Koutsoukos]{said2024neurographbenchmarksgraphmachine}
Anwar Said, Roza~G. Bayrak, Tyler Derr, Mudassir Shabbir, Daniel Moyer, Catie
  Chang, and Xenofon Koutsoukos.
\newblock {NeuroGraph}: Benchmarks for graph machine learning in brain
  connectomics.
\newblock In \emph{Advances in Neural Information Processing Systems},
  volume~36, pages 6509--6531, 2023.

\bibitem[Hamilton et~al.(2017)Hamilton, Ying, and
  Leskovec]{hamilton2018inductiverepresentationlearninglarge}
William~L. Hamilton, Rex Ying, and Jure Leskovec.
\newblock Inductive representation learning on large graphs.
\newblock In \emph{Advances in Neural Information Processing Systems},
  volume~30, 2017.

\bibitem[Kipf and
  Welling(2017)]{kipf2017semisupervisedclassificationgraphconvolutional}
Thomas~N. Kipf and Max Welling.
\newblock Semi-supervised classification with graph convolutional networks.
\newblock In \emph{International Conference on Learning Representations}, 2017.

\bibitem[Veličković et~al.(2018)Veličković, Cucurull, Casanova, Romero,
  Liò, and Bengio]{velickovic2018graphattentionnetworks}
Petar Veličković, Guillem Cucurull, Arantxa Casanova, Adriana Romero, Pietro
  Liò, and Yoshua Bengio.
\newblock Graph attention networks.
\newblock In \emph{International Conference on Learning Representations}, 2018.

\bibitem[Han et~al.(2026)Han, Su, He, Zhan, Plis, Calhoun, and
  Yang]{han2025rethinkingfunctionalbrainconnectome}
Keqi Han, Yao Su, Lifang He, Liang Zhan, Sergey Plis, Vince Calhoun, and Carl
  Yang.
\newblock Rethinking functional brain connectome analysis: Do graph deep
  learning models help?
\newblock \emph{npj Artificial Intelligence}, 2\penalty0 (1):\penalty0 19,
  2026.

\bibitem[Yang et~al.(2025)Yang, Liu, Zhuo, Jin, Wang, Wang, and
  Cao]{yang2025do}
Liang Yang, Yuwei Liu, Jiaming Zhuo, Di~Jin, Chuan Wang, Zhen Wang, and
  Xiaochun Cao.
\newblock Do we really need message passing in brain network modeling?
\newblock In \emph{Proceedings of the 42nd International Conference on Machine
  Learning}, volume 267 of \emph{Proceedings of Machine Learning Research},
  pages 70873--70887. PMLR, 2025.
\newblock URL \url{https://openreview.net/forum?id=KRosBwvhDx}.

\bibitem[Lara-Rangel and
  Heinbaugh(2025)]{lararangel2025limitsapplyinggraphtransformers}
Jose Lara-Rangel and Clare Heinbaugh.
\newblock On the limits of applying graph transformers for brain connectome
  classification.
\newblock In \emph{Proceedings on ``I Can't Believe It's Not Better: Challenges
  in Applied Deep Learning'' at ICLR 2025 Workshops}, volume 296 of
  \emph{Proceedings of Machine Learning Research}, pages 47--55. PMLR, 2025.

\bibitem[Li et~al.(2018)Li, Han, and
  Wu]{li2018deeperinsightsgraphconvolutional}
Qimai Li, Zhichao Han, and Xiao-Ming Wu.
\newblock Deeper insights into graph convolutional networks for semi-supervised
  learning.
\newblock In \emph{Proceedings of the AAAI Conference on Artificial
  Intelligence}, volume~32, pages 3538--3545, 2018.

\bibitem[Keriven(2022)]{keriven2022littlemuchtheoreticalanalysis}
Nicolas Keriven.
\newblock Not too little, not too much: a theoretical analysis of graph
  (over)smoothing.
\newblock In \emph{Advances in Neural Information Processing Systems},
  volume~35, pages 2268--2281, 2022.

\bibitem[Chen et~al.(2025)Chen, Schmidt-Hieber, Donnat, and
  Klopp]{chen2025understandingeffectgcnconvolutions}
Juntong Chen, Johannes Schmidt-Hieber, Claire Donnat, and Olga Klopp.
\newblock Understanding the effect of {GCN} convolutions in regression tasks.
\newblock In \emph{Proceedings of The 28th International Conference on
  Artificial Intelligence and Statistics}, volume 258 of \emph{Proceedings of
  Machine Learning Research}, pages 4573--4581. PMLR, 2025.

\bibitem[Rusch et~al.(2023)Rusch, Bronstein, and
  Mishra]{rusch2023surveyoversmoothinggraphneural}
T.~Konstantin Rusch, Michael~M. Bronstein, and Siddhartha Mishra.
\newblock A survey on oversmoothing in graph neural networks, 2023.
\newblock URL \url{https://arxiv.org/abs/2303.10993}.

\bibitem[Calhoun et~al.(2001)Calhoun, Adali, Pearlson, and
  Pekar]{calhoun2001method}
Vince~D. Calhoun, T{\"u}lay Adali, Godfrey~D. Pearlson, and James~J. Pekar.
\newblock A method for making group inferences from functional {MRI} data using
  independent component analysis.
\newblock \emph{Human Brain Mapping}, 14\penalty0 (3):\penalty0 140--151, 2001.
\newblock \doi{10.1002/hbm.1048}.

\bibitem[Beckmann and Smith(2004)]{beckmann2004probabilistic}
Christian~F. Beckmann and Stephen~M. Smith.
\newblock Probabilistic independent component analysis for functional magnetic
  resonance imaging.
\newblock \emph{IEEE Transactions on Medical Imaging}, 23\penalty0
  (2):\penalty0 137--152, 2004.
\newblock \doi{10.1109/TMI.2003.822821}.

\bibitem[Beckmann et~al.(2005)Beckmann, DeLuca, Devlin, and
  Smith]{beckmann2005investigations}
Christian~F. Beckmann, Mercedes DeLuca, Joseph~T. Devlin, and Stephen~M. Smith.
\newblock Investigations into resting-state connectivity using independent
  component analysis.
\newblock \emph{Philosophical Transactions of the Royal Society B: Biological
  Sciences}, 360\penalty0 (1457):\penalty0 1001--1013, 2005.
\newblock \doi{10.1098/rstb.2005.1634}.

\bibitem[Li et~al.(2021)Li, Zhou, Dvornek, Zhang, Gao, Zhuang, Scheinost,
  Staib, Ventola, and Duncan]{li2021braingnn}
Xiaoxiao Li, Yuan Zhou, Nicha Dvornek, Muhan Zhang, Siyuan Gao, Juntang Zhuang,
  Dustin Scheinost, Lawrence~H Staib, Pamela Ventola, and James~S Duncan.
\newblock {BrainGNN}: Interpretable brain graph neural network for {fMRI}
  analysis.
\newblock \emph{Medical image analysis}, 74:\penalty0 102233, 2021.

\bibitem[Febrinanto et~al.(2023)Febrinanto, Liu, and
  Xia]{febrinanto2023balanced}
Falih~Gozi Febrinanto, Mujie Liu, and Feng Xia.
\newblock Balanced graph structure information for brain disease detection.
\newblock In \emph{Pacific Rim Knowledge Acquisition Workshop}, pages 134--143.
  Springer, 2023.

\bibitem[Xu et~al.(2025{\natexlab{a}})Xu, He, Tang, Li, Lan, Dong, Ke, and
  Feng]{xu2025brainprompt}
Jiaxing Xu, Kai He, Yue Tang, Wei Li, Mengcheng Lan, Xia Dong, Yiping Ke, and
  Mengling Feng.
\newblock {BrainPrompt}: Multi-level brain prompt enhancement for neurological
  condition identification.
\newblock In \emph{International Conference on Medical Image Computing and
  Computer-Assisted Intervention}, pages 172--182. Springer,
  2025{\natexlab{a}}.

\bibitem[Xu et~al.(2025{\natexlab{b}})Xu, Chen, Dong, Lan, Huang, Bian, Cheng,
  and Ke]{xu2025brainood}
Jiaxing Xu, Yongqiang Chen, Xia Dong, Mengcheng Lan, Tiancheng Huang, Qingtian
  Bian, James Cheng, and Yiping Ke.
\newblock {BrainOOD}: Out-of-distribution generalizable brain network analysis.
\newblock In \emph{International Conference on Learning Representations},
  2025{\natexlab{b}}.

\bibitem[Battaglia et~al.(2018)Battaglia, Hamrick, Bapst, Sanchez-Gonzalez,
  Zambaldi, Malinowski, Tacchetti, Raposo, Santoro, Faulkner, Gulcehre, Song,
  Ballard, Gilmer, Dahl, Vaswani, Allen, Nash, Langston, Dyer, Heess, Wierstra,
  Kohli, Botvinick, Vinyals, Li, and
  Pascanu]{battaglia2018relationalinductivebiasesdeep}
Peter~W. Battaglia, Jessica~B. Hamrick, Victor Bapst, Alvaro Sanchez-Gonzalez,
  Vinicius Zambaldi, Mateusz Malinowski, Andrea Tacchetti, David Raposo, Adam
  Santoro, Ryan Faulkner, Caglar Gulcehre, Francis Song, Andrew Ballard, Justin
  Gilmer, George Dahl, Ashish Vaswani, Kelsey Allen, Charles Nash, Victoria
  Langston, Chris Dyer, Nicolas Heess, Daan Wierstra, Pushmeet Kohli, Matt
  Botvinick, Oriol Vinyals, Yujia Li, and Razvan Pascanu.
\newblock Relational inductive biases, deep learning, and graph networks, 2018.
\newblock URL \url{https://arxiv.org/abs/1806.01261}.

\bibitem[Wu et~al.(2019)Wu, Zhang, Souza, Fifty, Yu, and
  Weinberger]{wu2019simplifying}
Felix Wu, Tianyi Zhang, Amauri Holanda~de Souza, Christopher Fifty, Tao Yu, and
  Kilian~Q. Weinberger.
\newblock Simplifying graph convolutional networks.
\newblock In \emph{Proceedings of the 36th International Conference on Machine
  Learning}, pages 6861--6871, 2019.

\bibitem[NT and Maehara(2019)]{nt2019revisiting}
Hoang NT and Takanori Maehara.
\newblock Revisiting graph neural networks: All we have is low-pass filters.
\newblock \emph{arXiv preprint arXiv:1905.09550}, 2019.

\bibitem[He et~al.(2020)He, Deng, Wang, Li, Zhang, and Wang]{he2020lightgcn}
Xiangnan He, Kuan Deng, Xiang Wang, Yan Li, Yongdong Zhang, and Meng Wang.
\newblock {LightGCN}: Simplifying and powering graph convolution network for
  recommendation.
\newblock In \emph{Proceedings of the 43rd International ACM SIGIR Conference
  on Research and Development in Information Retrieval}, pages 639--648, 2020.

\bibitem[Loh and Wainwright(2011)]{loh2011high}
Po-Ling Loh and Martin~J Wainwright.
\newblock High-dimensional regression with noisy and missing data: Provable
  guarantees with non-convexity.
\newblock In \emph{Advances in Neural Information Processing Systems},
  volume~24, pages 2726--2734, 2011.

\bibitem[Datta and Zou(2017)]{datta2017cocolasso}
Abhirup Datta and Hui Zou.
\newblock {CoCoLasso} for high-dimensional error-in-variables regression.
\newblock \emph{Annals of Statistics}, 45\penalty0 (6):\penalty0 2400--2426,
  2017.

\bibitem[Stefanski and Carroll(1987)]{stefanski1987conditional}
Leonard~A Stefanski and Raymond~J Carroll.
\newblock Conditional scores and optimal scores for generalized linear
  measurement-error models.
\newblock \emph{Biometrika}, 74\penalty0 (4):\penalty0 703--716, 1987.

\bibitem[Chen et~al.(2011)Chen, Hong, and Nekipelov]{chen2011nonlinear}
Xiaohong Chen, Han Hong, and Denis Nekipelov.
\newblock Nonlinear models of measurement errors.
\newblock \emph{Journal of Economic Literature}, 49\penalty0 (4):\penalty0
  901--937, 2011.

\bibitem[Schennach(2013)]{schennach2013measurement}
Susanne~M Schennach.
\newblock Measurement error in nonlinear models--a review.
\newblock In Daron Acemoglu, Manuel Arellano, and Eddie Dekel, editors,
  \emph{Advances in Economics and Econometrics: Tenth World Congress},
  volume~3, pages 296--337. Cambridge University Press, 2013.

\bibitem[Hu et~al.(2022)Hu, Ke, and Liu]{hu2022measurement}
Zhirui Hu, Zheng~Tracy Ke, and Jun~S Liu.
\newblock Measurement error models: from nonparametric methods to deep neural
  networks.
\newblock \emph{Statistical Science}, 37\penalty0 (4):\penalty0 473--493, 2022.

\bibitem[Airoldi et~al.(2008)Airoldi, Blei, Fienberg, and
  Xing]{airoldi2007mixedmembershipstochasticblockmodels}
Edoardo~M Airoldi, David~M Blei, Stephen~E Fienberg, and Eric~P Xing.
\newblock Mixed membership stochastic blockmodels.
\newblock \emph{Journal of Machine Learning Research}, 9:\penalty0 1981--2014,
  2008.

\bibitem[Wishart(1928)]{1626a8bf-8479-36bf-b395-22e4def81091}
John Wishart.
\newblock The generalised product moment distribution in samples from a normal
  multivariate population.
\newblock \emph{Biometrika}, 20A\penalty0 (1/2):\penalty0 32--52, 1928.
\newblock ISSN 00063444, 14643510.
\newblock URL \url{http://www.jstor.org/stable/2331939}.

\bibitem[Magnus and Neudecker(1979)]{10.1214/aos/1176344621}
Jan~R. Magnus and H.~Neudecker.
\newblock {The Commutation Matrix: Some Properties and Applications}.
\newblock \emph{The Annals of Statistics}, 7\penalty0 (2):\penalty0 381 -- 394,
  1979.
\newblock \doi{10.1214/aos/1176344621}.
\newblock URL \url{https://doi.org/10.1214/aos/1176344621}.

\bibitem[Worsley et~al.(2002)Worsley, Liao, Aston, Petre, Duncan, Morales, and
  Evans]{Worsleyetal2002}
K.~J. Worsley, C.~H. Liao, J.~Aston, V.~Petre, G.~H. Duncan, F.~Morales, and
  A.~C. Evans.
\newblock A general statistical analysis for {fMRI} data.
\newblock \emph{NeuroImage}, 15\penalty0 (1):\penalty0 1--15, 2002.
\newblock \doi{10.1006/nimg.2001.0933}.

\bibitem[{The ADHD-200 Consortium}(2012)]{ADHD200Consortium2012}
{The ADHD-200 Consortium}.
\newblock The {ADHD-200} consortium: a model to advance the translational
  potential of neuroimaging in clinical neuroscience.
\newblock \emph{Frontiers in Systems Neuroscience}, 6:\penalty0 62, 2012.
\newblock \doi{10.3389/fnsys.2012.00062}.

\bibitem[Bellec et~al.(2017)Bellec, Chu, Chouinard-Decorte, Benhajali,
  Margulies, and Craddock]{bellec2017neuro}
Pierre Bellec, Carlton Chu, Fran{\c{c}}ois Chouinard-Decorte, Yassine
  Benhajali, Daniel~S Margulies, and R~Cameron Craddock.
\newblock The {Neuro Bureau} {ADHD-200} preprocessed repository.
\newblock \emph{NeuroImage}, 144:\penalty0 275--286, 2017.
\newblock \doi{10.1016/j.neuroimage.2016.06.034}.

\bibitem[Craddock et~al.(2012)Craddock, James, Holtzheimer, Hu, and
  Mayberg]{Craddock2012}
R.~C. Craddock, G.~A. James, P.~E. Holtzheimer, X.~P. Hu, and H.~S. Mayberg.
\newblock A whole brain {fMRI} atlas generated via spatially constrained
  spectral clustering.
\newblock \emph{Human Brain Mapping}, 33\penalty0 (8):\penalty0 1914--1928,
  2012.
\newblock \doi{10.1002/hbm.21333}.

\bibitem[Di~Martino et~al.(2014)Di~Martino, Yan, Li, Denio, Castellanos,
  Alaerts, Anderson, Assaf, Bookheimer, Dapretto, et~al.]{dimartino2014autism}
A.~Di~Martino, C.-G. Yan, Q.~Li, E.~Denio, F.~X. Castellanos, K.~Alaerts, J.~S.
  Anderson, M.~Assaf, S.~Y. Bookheimer, M.~Dapretto, et~al.
\newblock The autism brain imaging data exchange: towards a large-scale
  evaluation of the intrinsic brain architecture in autism.
\newblock \emph{Molecular Psychiatry}, 19\penalty0 (6):\penalty0 659--667,
  2014.
\newblock \doi{10.1038/mp.2013.78}.
\newblock URL \url{https://doi.org/10.1038/mp.2013.78}.

\bibitem[Craddock et~al.(2013{\natexlab{a}})Craddock, Benhajali, Chu,
  Chouinard, Evans, Jakab, Khundrakpam, Lewis, Li, Milham,
  et~al.]{craddock2013neuro}
Cameron Craddock, Yassine Benhajali, Carlton Chu, Francois Chouinard, Alan
  Evans, Andr{\'a}s Jakab, Budhachandra~Singh Khundrakpam, John~David Lewis,
  Qingyang Li, Michael Milham, et~al.
\newblock The {Neuro Bureau} preprocessing initiative: open sharing of
  preprocessed neuroimaging data and derivatives.
\newblock In \emph{Frontiers in Neuroinformatics Conference Abstract:
  Neuroinformatics 2013}. Frontiers Media SA, 2013{\natexlab{a}}.
\newblock \doi{10.3389/conf.fninf.2013.09.00041}.
\newblock URL
  \url{https://www.frontiersin.org/10.3389/conf.fninf.2013.09.00041/event_abstract}.

\bibitem[Craddock et~al.(2013{\natexlab{b}})Craddock, Sikka, Cheung, Khanuja,
  Ghosh, Yan, Li, Lurko, Vogelstein, Burns, Colcombe, Mennes, Kelly,
  Di~Martino, Castellanos, and Milham]{Craddocketal2013_cpac}
Cameron Craddock, Sharad Sikka, Brian Cheung, Ranjeet Khanuja, Satrajit~S
  Ghosh, Chaogan Yan, Qingyang Li, V~Lurko, Joshua Vogelstein, Randal Burns,
  Stanley Colcombe, Maarten Mennes, Clare Kelly, Adriana Di~Martino,
  Francisco~Xavier Castellanos, and Michael Milham.
\newblock Towards automated analysis of connectomes: The configurable pipeline
  for the analysis of connectomes (c-pac).
\newblock In \emph{Frontiers in Neuroinformatics Conference Abstract:
  Neuroinformatics 2013}. Frontiers Media SA, 2013{\natexlab{b}}.
\newblock \doi{10.3389/conf.fninf.2013.09.00042}.
\newblock URL
  \url{https://www.frontiersin.org/10.3389/conf.fninf.2013.09.00042/event_abstract}.

\bibitem[Abraham et~al.(2014)Abraham, Pedregosa, Eickenberg, Gervais, Mueller,
  Kossaifi, Gramfort, Thirion, and Varoquaux]{Abrahametal2014}
Alexandre Abraham, Fabian Pedregosa, Michael Eickenberg, Philippe Gervais,
  Andreas Mueller, Jean Kossaifi, Alexandre Gramfort, Bertrand Thirion, and
  Ga{\"e}l Varoquaux.
\newblock Machine learning for neuroimaging with scikit-learn.
\newblock \emph{Frontiers in Neuroinformatics}, 8:\penalty0 14, 2014.
\newblock \doi{10.3389/fninf.2014.00014}.

\bibitem[Tagliazucchi et~al.(2012)Tagliazucchi, von Wegner, Morzelewski,
  Borisov, Jahnke, and Laufs]{Tagliazucchietal2012}
E.~Tagliazucchi, F.~von Wegner, A.~Morzelewski, S.~Borisov, K.~Jahnke, and
  H.~Laufs.
\newblock Automatic sleep staging using {fMRI} functional connectivity data.
\newblock \emph{NeuroImage}, 63\penalty0 (1):\penalty0 63--72, 2012.
\newblock \doi{10.1016/j.neuroimage.2012.06.036}.

\bibitem[Gordon et~al.(2016)Gordon, Laumann, Adeyemo, Huckins, Kelley, and
  Petersen]{Gordon2016Generation}
Evan~M. Gordon, Timothy~O. Laumann, Babatunde Adeyemo, Jeremy~F. Huckins,
  William~M. Kelley, and Steven~E. Petersen.
\newblock Generation and evaluation of a cortical area parcellation from
  resting-state correlations.
\newblock \emph{Cerebral Cortex}, 26\penalty0 (1):\penalty0 288--303, 2016.
\newblock \doi{10.1093/cercor/bhu239}.

\end{thebibliography}
\clearpage

\appendix

\section{Full theoretical analysis}
\subsection{Technical preliminaries}\label{app:general}
We state important definitions and technical preliminaries that we build on in the results that follow. 

\begin{lemma}[Wishart noise moments]
\label{lem:wishart}
Let $\Sigma \in \mathbb{R}^{p \times p}$ be a positive-definite covariance
matrix and let $m\hat{\Sigma} \sim \mathcal{W}_p(m, \Sigma)$~\citep{1626a8bf-8479-36bf-b395-22e4def81091}. Then
$\hat{\Sigma} = \Sigma + H$ with:
\begin{enumerate}
\item $\mathbb{E}_{\eta}[H_{jk} \mid \Sigma] = 0$,
\item $\mathrm{Cov}_{\eta}(H_{jk},\, H_{lm} \mid \Sigma)
      = \dfrac{\Sigma_{jl}\,\Sigma_{km} + \Sigma_{jm}\,\Sigma_{kl}}{m}$.
\end{enumerate}
\end{lemma}

Next, we show that $\hat{A}$ and $A^\perp$ have the same marginal graph distribution. Their difference is the dependence between $\hat{A}$ and the feature noise; see Proposition~\ref{prop:double_dipping_window}, where we isolate the remaining difference, namely their joint distribution with the feature noise.
\begin{lemma}[Marginal equivalence]
\label{lem:marginal}
Let $\hat{A}^{(i)} = \mathbbm 1\{\hat{\Sigma}^{(i)} > \delta\}$ and
$A^{\perp (i)} = \mathbbm 1\{\widehat\Sigma^{\perp(i)} > \delta\}$, where
$\widehat\Sigma^{\perp(i)} = \Sigma^{(i)} + H^{\perp(i)}$ with
$H^{\perp(i)} \indep H^{(i)} \mid \Sigma^{(i)}$ and
$H^{\perp(i)} \stackrel{d}{=} H^{(i)} \mid \Sigma^{(i)}$.
Then:
\begin{enumerate}
\item $\hat{A}^{(i)}$ and $A^{\perp (i)}$ have the same conditional
      distribution given $\Sigma^{(i)}$.
\item $A^{\perp (i)} \indep H^{(i)} \mid \Sigma^{(i)}$,
      whereas $\hat{A}^{(i)}$ depends on $H^{(i)}$ through the shared
      thresholding.
\end{enumerate}
\end{lemma}

Consequently, all marginal quantities (edge probabilities, expected
neighbor counts, expected degree) are identical for $\hat{A}$ and
$A^\perp$. The two differ only in their joint distribution with
the feature noise $\eta = \mathrm{vec}(H)$.

With $\operatorname{vec}$ stacking columns, $\operatorname{vec}(ABC)=(C^\top\otimes A)\operatorname{vec}(B)$; in particular, $\operatorname{vec}(R\widehat\Sigma)=(I_p\otimes R)\operatorname{vec}(\widehat\Sigma)$.
We also define the commutation matrix as it plays a crucial role when working with vectorized covariance matrices. 
\begin{definition}[Commutation matrix]
\label{def:commutation}
The commutation matrix $K_{p,p} \in \mathbb{R}^{p^2 \times p^2}$ is the
unique permutation matrix satisfying
$K_{p,p}\,\mathrm{vec}(A) = \mathrm{vec}(A^\top)$ for every
$A \in \mathbb{R}^{p \times p}$ \cite{10.1214/aos/1176344621}.

Writing $\Psi(\Sigma) := (I_{p^2}+K_{p,p})(\Sigma\otimes\Sigma)$, from Lemma~\ref{lem:wishart}, we note:
\begin{enumerate}
\item $K_{p,p}$ is symmetric and involutory (i.e.\ $K_{p,p}^2 = I_{p^2}$), and
$K_{p,p}(B\otimes C) = (C\otimes B)K_{p,p}$ for all $B,C \in \mathbb{R}^{p\times p}$
\cite{10.1214/aos/1176344621}; in particular $K_{p,p}$ commutes with
$\Sigma\otimes\Sigma$, so $\Psi(\Sigma)$ is symmetric.
\item $\Psi(\Sigma) \succeq 0$, since $T^{-1}\Psi(\Sigma)$ is a conditional
covariance matrix (Lemma~\ref{lem:wishart}).
\item $\Psi(\Sigma)\,\mathrm{vec}(W) = 0$ for every antisymmetric
$W = -W^\top$: indeed $\Sigma W\Sigma$ is again antisymmetric, and
$(I_{p^2}+K_{p,p})\,\mathrm{vec}(M) = \mathrm{vec}(M) + \mathrm{vec}(M^\top) = 0$
for antisymmetric $M$. Hence $\Psi(\Sigma)$ is singular, with
$\operatorname{rank}\Psi(\Sigma) \le p(p+1)/2$.
\end{enumerate}
\end{definition}

\subsection{Statistical framework}\label{app:stats_framework}
For each subject $i=1,\ldots,n$, the centered time series $Z^{(i)}\in\R^{T\times p}$ has rows
$Z^{(i)}_t\overset{\mathrm{i.i.d.}}{\sim}N(0,\Sigma^{(i)})$, $t=1,\ldots,T$, and we observe the empirical covariance
\[
    \widehat\Sigma^{(i)}=\frac1T\sum_{t=1}^T Z^{(i)}_tZ^{(i)\top}_t=\Sigma^{(i)}+H^{(i)} .
\]
The analysis is in the population regime, $n\to\infty$. Let
\[
    x:=\vecop(\Sigma),
    \qquad
    \widehat x:=\vecop(\widehat\Sigma)=x+\eta,
    \qquad
    \eta:=\vecop(H).
\]
By Lemma~\ref{lem:wishart},
\[
    \E_{\eta}[\eta\mid \Sigma]=0,
    \qquad
    \Cov_{\eta}(\eta\mid \Sigma)=T^{-1}\Psi(\Sigma),
    \qquad
    \Psi(\Sigma):=(I_{p^2}+K_{p,p})(\Sigma\otimes\Sigma).
\]
The demeaned estimator has $T-1$ degrees of freedom; every statement below holds with $T-1$ in place of $T$. The target depends on the latent connectome, $Y=f^\star(\Sigma)+\varepsilon$, with $\varepsilon$ independent of $(\Sigma,H)$ and of any randomness used to build a graph, and $\E_{\Sigma}\|\Sigma\|_F^4<\infty$. We restrict $f^\star$ to the linear class $\mathcal F=\{x\mapsto x^\top\theta\}$, so that
\[
    Y=h(\Sigma)+\varepsilon,
    \qquad
    h(\Sigma):=x^\top\beta^\star=\inner{\Sigma}{B^\star},
    \qquad
    \vecop(B^\star)=\beta^\star,
    \qquad
    \varepsilon\sim N(0,\sigma_\varepsilon^2),
\]
with $B^\star$ symmetric without loss of generality, since $\inner{\Sigma}{B^\star}$ only sees the symmetric part of $B^\star$.
Since $\widehat\Sigma$ is symmetric, the second-moment matrix of $\widehat x$ and the Gram matrices below are singular by Definition~\ref{def:commutation}; all inverses are Moore--Penrose pseudo-inverses.
For the raw feature map $\widehat x$ we write
\[
    S_0:=\E_{\Sigma}[xx^\top],
    \qquad
    \Gamma_0:=\E_{\Sigma}[\Psi(\Sigma)],
    \qquad
    c_0:=\E_{\Sigma}[xh(\Sigma)]=S_0\beta^\star.
\]
Thus
\[
    \mse_{\mathrm{raw}}(T)
    =
    \E_{\Sigma}[h(\Sigma)^2]
    -c_0^\top(S_0+T^{-1}\Gamma_0)^{\dagger}c_0
    +\sigma_\varepsilon^2,
\]
We now state and prove Remark~\ref{thm:raw1}, whose $S,\Gamma$ are the $S_0,\Gamma_0$ above and whose $\mathrm{MSE}_{\mathrm{raw}}(T)$ is the display above written through $\Lambda(T)$.

\begin{remark}[Raw Feature Map]\label{thm:raw1}
Let \(y_i=x_i^{\top}\beta^\star+\varepsilon_i\), where
\(\Sigma^{(i)}\sim P_\Sigma\) i.i.d. across subjects,
\(x_i=\mathrm{vec}(\Sigma^{(i)})\), \(\varepsilon_i\overset{\mathrm{i.i.d.}}{\sim}
\mathcal N(0,\sigma_\varepsilon^2)\) independent of \((\Sigma^{(i)},\eta_i)\), and
\(\hat x_i=x_i+\eta_i\) with \(\mathbb E_{\eta_i}[\eta_i\mid \Sigma^{(i)}]=0\). Let
\[
\mathrm{Cov}_{\eta_i}(\eta_i\mid\Sigma^{(i)})=\frac{1}{T}\Psi^{(i)},\qquad
\Psi^{(i)}=(I_{p^2}+K_{p,p})(\Sigma^{(i)}\otimes\Sigma^{(i)}),
\]
where \(K_{p,p}\) is the commutation matrix (Definition~\ref{def:commutation}).
Write \(S=\mathbb E_{\Sigma}[xx^{\top}]\) and \(\Gamma=\mathbb E_{\Sigma}[\Psi]\).
Since the features are vectorized symmetric matrices, \(S+\frac{1}{T}\Gamma\) is
singular by Appendix~\ref{app:general}, so inverses below are
Moore--Penrose pseudo-inverses. The population raw-feature OLS coefficient is
\[
\beta_{\mathrm{raw}}^*(T)=\Lambda(T)\beta^\star,\qquad
\Lambda(T):=(S+\tfrac{1}{T}\Gamma)^{\dagger}S,
\]
so, as \(T\to\infty\), provided the support of \(P_\Sigma\) spans the symmetric
matrices, \(\beta_{\mathrm{raw}}^*(T)\to S^\dagger S\,\beta^\star\), the
orthogonal projection of \(\beta^\star\) onto \(\mathrm{range}(S)\); the limiting
predictor nevertheless equals \(x^{\top}\beta^\star\) almost surely.
Its population test error is
\[
\mathrm{MSE}_{\mathrm{raw}}(T)
=
\beta^{\star\top}(I-\Lambda(T))^\top S(I-\Lambda(T))\beta^\star
+
\frac{1}{T}\beta^{\star\top}\Lambda(T)^\top\Gamma\Lambda(T)\beta^\star
+\sigma_\varepsilon^2 .
\]
\end{remark}

\begin{proof}
For any coefficient $\gamma$, conditional centering of $\eta$
and independence of $\varepsilon$ make the cross terms vanish:
\[
\begin{aligned}
    \E_{\Sigma,\eta,\varepsilon}(y-\hat x^\top\gamma)^2
    &= \E_{\Sigma,\eta,\varepsilon}\{x^\top(\beta^\star-\gamma)
           -\eta^\top\gamma+\varepsilon\}^2\\
    &= (\beta^\star-\gamma)^\top S(\beta^\star-\gamma)
       +T^{-1}\gamma^\top\Gamma\gamma
       +\sigma_\varepsilon^2.
\end{aligned}
\]
Minimizing this quadratic gives
$(S+T^{-1}\Gamma)\gamma=S\beta^\star$, whose minimum-norm
solution is $\beta_{\mathrm{raw}}^*(T)=\Lambda(T)\beta^\star$.
Substituting into the same expansion gives the stated MSE.

Under the spanning assumption, $S$ is invertible on the
symmetric subspace, while both $S$ and $\Gamma$ vanish on
its orthogonal complement. Thus $\Lambda(T)\to S^\dagger S$.
Since $x$ belongs to that subspace,
$x^\top S^\dagger S\beta^\star=x^\top\beta^\star$:
the limiting predictor recovers the noiseless target.
\end{proof}
A linear graph convolution replaces $\widehat\Sigma$ by $R(A)\widehat\Sigma$, where $A\in\{0,1\}^{p\times p}$ is a symmetric adjacency matrix and $R(A)$ its message-passing operator: the layer $X'=\sigma(R(A)XW)$ with the weights and the activation removed. The smoothed feature map is
\[
    \widetilde x_A:=\vecop(R(A)\widehat\Sigma)=\cR_A\widehat x,
    \qquad
    \cR_A:=I_p\otimes R(A),
\]
and, for a graph $A$, $\mse_A(T):=\min_{\gamma}\E_{\Sigma,\eta,\varepsilon,A}[(Y-\gamma^\top\widetilde x_A)^2]$ is the optimized population risk, the object compared throughout; $\mse_{\mathrm{raw}}(T)$ above is its value for the raw features $\widehat x$.
For a binary adjacency matrix \(A\), the SAGE mean-aggregation operator \(R=R(A)\) is
\[
    \tildA=A+I_p,
    \qquad
    \tildD=\diag(\tildA\ones),
    \qquad
    R=\tildD^{-1}\tildA.
\]
We compare four graphs, thresholded entrywise off the diagonal at a fixed $\delta$:
\begin{enumerate}[leftmargin=1.5em, itemsep=0pt, topsep=0pt]
    \item the population graph $A^{\text{pop}
    }=\mathbbm 1\{\Sigma_0>\delta\}$, with $\Sigma_0=\E_\Sigma[\Sigma]$;
    \item the oracle graph $A^{\text{oracle}}=\mathbbm 1\{\Sigma>\delta\}$;
    \item the same-sample graph $\widehat A=\mathbbm 1\{\widehat\Sigma>\delta\}$;
    \item the independent graph $A^\perp=\mathbbm 1\{\widehat\Sigma^\perp>\delta\}$, $\widehat\Sigma^\perp=\Sigma+H^\perp$, with $H^\perp\indep H\mid\Sigma$ and $H^\perp\overset{d}{=}H\mid\Sigma$.
\end{enumerate}

\subsection{General errors-in-variables analysis}\label{app:theory_eiv}

Predicting from the smoothed features $\widetilde x_A=\cR_A\widehat x$ rather than from $\widehat x$ is still an errors-in-variables regression, now with the operator $\cR_A$ sitting between the noise and the coefficient. We characterize its population MSE without assumptions on $\Sigma$, $B^\star$ or $A$, distinguishing the case where the graph is independent of the feature noise from the case where graph and features are derived from the same sample. The fixed-operator case, in which the graph does not vary across subjects, is treated separately: there message passing cannot improve on raw features.

\begin{proposition}[Population MSE under message passing]\label{prop:eiv-message-passing}
For a graph $A$, let $R(A) \in \mathbb{R}^{p \times p}$ denote its convolution operator ($R(A) = \widetilde D^{-1}\widetilde A$ for GraphSAGE), $\mathcal{R}_A := I_p \otimes R(A)$ the corresponding vectorized operator, so that $\operatorname{vec}(R(A)\widehat\Sigma) = \mathcal{R}_A\widehat x$, and $\widetilde{x}_A := \mathcal{R}_A(x + \eta)$ the smoothed feature map, where $h(\Sigma) := \mathrm{vec}(\Sigma)^\top\beta^\star$ is the noiseless regression target and $\Psi$ is as in Remark~\ref{thm:raw1}.

\textbf{(a) Conditionally independent graph.} Suppose $A \indep H \mid \Sigma$, as for $A^{\mathrm{pop}}$, $A^{\text{oracle}}$, and $A^\perp$, and $\varepsilon\indep(\Sigma,\eta,A)$. Define
\[
    S_A := \mathbb{E}_{\Sigma, A}[\mathcal{R}_A xx^\top\mathcal{R}_A^\top], \qquad \Gamma_A := \mathbb{E}_{\Sigma, A}[\mathcal{R}_A\Psi\mathcal{R}_A^\top], \qquad c_A := \mathbb{E}_{\Sigma, A}[\mathcal{R}_A x\,h(\Sigma)].
\]
Then the population OLS coefficient and test MSE are
\[
    \beta^*_A(T) = (S_A + T^{-1}\Gamma_A)^{\dagger}c_A, \qquad \mathrm{MSE}_A(T) = \mathbb{E}_{\Sigma}[h(\Sigma)^2] - c_A^\top(S_A + T^{-1}\Gamma_A)^{\dagger}c_A + \sigma_\varepsilon^2.
\]

\textbf{(b) Same-sample graph.} Suppose $A = \widehat A(\widehat\Sigma)$ is the same-sample graph, so that $\mathcal{R}_{\widehat A}$ is $\widehat\Sigma$-measurable and no longer independent of $\eta$ given $\Sigma$, and $\varepsilon\indep(\Sigma,\eta)$. Define $M_{\widehat{A}} := \mathbb{E}_{\Sigma, \eta}[\widetilde{x}_{\widehat A}\,\widetilde{x}_{\widehat A}^\top],\,c_{\widehat{A}} := \mathbb{E}_{\Sigma, \eta}[\widetilde{x}_{\widehat A}\,h(\Sigma)].$
Then $\beta^*_{\widehat A}(T) = M_{\widehat{A}}^{\dagger} c_{\widehat{A}}$ and $\mathrm{MSE}_{\mathrm{same}}(T) = \mathbb{E}_{\Sigma}[h(\Sigma)^2] - c_{\widehat{A}}^\top M_{\widehat{A}}^{\dagger} c_{\widehat{A}} + \sigma_\varepsilon^2$, where
\[
\begin{aligned}
    M_{\widehat{A}} &= \mathbb{E}_{\Sigma, \eta}[\mathcal{R}_{\widehat A}\,xx^\top\mathcal{R}_{\widehat A}^\top]
    + \underbrace{\mathbb{E}_{\Sigma, \eta}[\mathcal{R}_{\widehat A}\,x\eta^\top\mathcal{R}_{\widehat A}^\top]
    + \mathbb{E}_{\Sigma, \eta}[\mathcal{R}_{\widehat A}\,\eta x^\top\mathcal{R}_{\widehat A}^\top]}_{\text{cross terms}}
    + \mathbb{E}_{\Sigma, \eta}[\mathcal{R}_{\widehat A}\,\eta\eta^\top\mathcal{R}_{\widehat A}^\top], \\
    c_{\widehat{A}} &= \mathbb{E}_{\Sigma, \eta}[\mathcal{R}_{\widehat A}\,xx^\top]\beta^\star
    + \mathbb{E}_{\Sigma, \eta}[\mathcal{R}_{\widehat A}\,\eta x^\top]\beta^\star .
\end{aligned}
\]
\end{proposition}

\begin{proof}[Proof of (a)]
Since $A\indep H\mid\Sigma$ and $\varepsilon\indep(\Sigma,\eta,A)$, we have $\E_{\eta}[\eta\mid\Sigma,A]=0$ and
$\Cov_{\eta}(\eta\mid\Sigma,A)=T^{-1}\Psi$. By the tower property,
\[
\E_{\Sigma,\eta,A}[\widetilde x_A\widetilde x_A^\top]=S_A+T^{-1}\Gamma_A,
\qquad
\E_{\Sigma,\eta,\varepsilon,A}[\widetilde x_A\,Y]=c_A,
\]and the normal equations
$(S_A+T^{-1}\Gamma_A)\gamma=c_A$ are solvable as in Remark~\ref{thm:raw1}, and
the minimal-norm solution $(S_A+T^{-1}\Gamma_A)^\dagger c_A$ yields
$\mse_A(T)$.
\end{proof}

\begin{proof}[Proof of (b)]
The expansions follow by linearity from
$\widetilde x_{\widehat A}=\cR_{\widehat A}(x+\eta)$, with $\E_{\Sigma,\eta,\varepsilon}[\widetilde x_{\widehat A}\,\varepsilon]=0$. No
term vanishes in the general case because  $\cR_{\widehat A}$ is
not conditionally independent of $\eta$ given $\Sigma$. The coefficient and MSE
follow as in Remark~\ref{thm:raw1}. If
$\cR_{\widehat A}\indep\eta\mid\Sigma$, then
$\E_{\eta}[\eta\mid\Sigma,\widehat A]=0$, so the cross terms and
$\E_{\Sigma,\eta}[\cR_{\widehat A}\,\eta x^\top]\beta^\star$ vanish, while the tower property
gives $\E_{\Sigma,\eta}[\cR_{\widehat A}\,\eta\eta^\top\cR_{\widehat A}^\top]
=T^{-1}\E_{\Sigma,\eta}[\cR_{\widehat A}\,\Psi\,\cR_{\widehat A}^\top]$, recovering case (a).
\end{proof}

For the same-sample graph, the operator depends on the same noise used in the
features. The previous cancellation no longer occurs.

\begin{proposition}[No linear gain from fixed message passing]\label{prop:no_linear_gain}
For any fixed \(\mathcal R\in\mathbb R^{p^2\times p^2}\),
\[
\mathcal F_{\mathrm{MP}}(\mathcal R):=\{x\mapsto \theta^\top\mathcal R x:\theta\in\mathbb R^{p^2}\}
\subseteq
\{x\mapsto \beta^\top x:\beta\in\mathbb R^{p^2}\}=:\mathcal F_{\mathrm{raw}},
\qquad
\theta^\top\mathcal R x=(\mathcal R^\top\theta)^\top x.
\]
Consequently, for any loss \(\ell\), including \(\ell(y,\hat y)=(y-\hat y)^2\):
$\inf_{\beta}\mathbb E_{Y,x}\,\ell(Y,\beta^\top x)
\le
\inf_{\theta}\mathbb E_{Y,x}\,\ell(Y,\theta^\top\mathcal R x).$
\end{proposition}

\begin{proof}
As $\theta$ ranges over $\mathbb R^{p^2}$, $\mathcal R^\top\theta$ ranges over
$\mathrm{range}(\mathcal R^\top)$, so
$\mathcal F_{\mathrm{MP}}(\mathcal R)=\{x\mapsto\beta^\top x:\beta\in
\mathrm{range}(\mathcal R^\top)\}\subseteq\mathcal F_{\mathrm{raw}}$, and the
infimum over a subset is at least the infimum over the set. For squared loss the
raw minimizers form the solution set of $\mathbb E_{x}[xx^\top]\beta=\mathbb E_{Y,x}[xY]$
(consistent, as in Remark~\ref{thm:raw1}); the inequality is strict iff this set
is disjoint from $\mathrm{range}(\mathcal R^\top)$.
\end{proof}

\begin{corollary}[Equality when the fixed operator is invertible]\label{cor:invertible_equality}
If \(\mathcal{R}\) is invertible, then
\[
    \mathcal F_{\mathrm{MP}}(\mathcal R)
    =
    \mathcal F_{\mathrm{raw}}.
\]
Hence the optimal unregularized linear prediction risks are equal:
\[
    \inf_{\beta}
    \mathbb E_{Y,x}\!\left[
    \left(Y-\beta^\top x\right)^2
    \right]
    =
    \inf_{\theta}
    \mathbb E_{Y,x}\!\left[
    \left(Y-\theta^\top \mathcal{R}x\right)^2
    \right].
\]
\end{corollary}
\begin{proof}
Proposition~\ref{prop:no_linear_gain} gives
\[
    \mathcal F_{\mathrm{MP}}(\mathcal R)
    \subseteq
    \mathcal F_{\mathrm{raw}}.
\]
If \(\mathcal{R}\) is invertible, then for every \(\beta\) we may choose
\[
    \theta = \mathcal{R}^{-\top}\beta,
\]
so that
\[
    \theta^\top \mathcal{R}x
    =
    \beta^\top x.
\]
Therefore
\[
    \mathcal F_{\mathrm{raw}}
    \subseteq
    \mathcal F_{\mathrm{MP}}(\mathcal R),
\]
and the two classes are identical.
\end{proof}

Beyond Proposition~\ref{prop:no_linear_gain}, neither case of Proposition~\ref{prop:eiv-message-passing} yields a conclusion without assumptions: the population risks depend on three ingredients the propositions leave free. We fix each under the latent community model of Appendix~\ref{app:latent} and study their interaction as $T$ varies.
\begin{enumerate}[leftmargin=1.5em, itemsep=0pt, topsep=0pt]
    \item \textbf{Community structure, $\Sigma$.} $S$, $\Gamma=\E_\Sigma[\Psi(\Sigma)]$ and $c$, hence every MSE above, are functionals of the law of $\Sigma$: its community organization and subject-level variability decide what a neighborhood average can remove (Appendix~\ref{app:lcm_structure}).
    \item \textbf{Predictive signal, $B^\star$.} The signal enters through $R(A)^\top B^\star$, so smoothing preserves it when $B^\star$ is nearly constant over the averaged neighborhoods and distorts it when the signal lies in differences between entries: community-smooth versus edge or contrast targets (Appendix~\ref{app:lcm_alignment}).
    \item \textbf{Graph and threshold, $A$ and $\delta$.} Which matrix is thresholded ($\Sigma$, $\widehat\Sigma^\perp$, $\widehat\Sigma$, $\Sigma_0$) fixes the coupling between graph and feature noise, the split of Proposition~\ref{prop:eiv-message-passing}(a)/(b); the threshold $\delta$ fixes the neighborhoods, so for graphs $\mathbbm 1\{\cdot>\delta\}$ the structure reduces to the position of $\delta$ relative to the between- and within-community levels of $\Sigma$ (Appendix~\ref{app:lcm_threshold}).
\end{enumerate}

\subsection{Latent community model}\label{app:latent}
In order to get more insight into the intractable problem defined in the previous section, we place assumptions on all three ingredients at once: $\Sigma$ is generated by a latent community model, $\delta$ is set relative to its community levels, and $B^\star$ is characterized by its alignment with the communities.

\subsubsection{Community structure}\label{app:lcm_structure}
We model the $p$ ROIs through $K\le\min\{p,d\}$ latent
communities~\citep{Bullmore2009}, motivated by low-dimensional
representations of brain activity~\citep{calhoun2001method,
beckmann2004probabilistic,beckmann2005investigations}.
Let $F\in\mathbb R^{p\times d}$ collect their latent loadings.
The structural parameter is the subject-level variability $\sigma$;
all other model parameters are held fixed.

For each subject $i$, the connectome is generated as follows.
\begin{enumerate}[leftmargin=1.5em, itemsep=0pt, topsep=0pt]
    \item \textbf{Node assignment:} each ROI $j$ belongs to
    $z_j\in\{1,\ldots,K\}$, with balanced communities
    $|C_k|=q=p/K\in\mathbb N$ and $C(j):=C_{z_j}$.
    \item \textbf{Community centroids:} the vectors
    $\mu_k\in\mathbb R^d$ are fixed across subjects, orthogonal,
    and satisfy $\|\mu_k\|^2=\phi^2d$.
    \item \textbf{Subject-level variability:}
    $\nu_j^{(i)}\sim\mathcal N(0,\sigma^2I_d)$ independently
    across ROIs and subjects.
    \item \textbf{Latent features:}
    $f_j^{(i)}=\mu_{z_j}+\nu_j^{(i)}$, with
    $f_j^{(i)\top}$ the $j$-th row of $F^{(i)}$.
    \item \textbf{Connectome:}
    $\Sigma^{(i)}=F^{(i)}F^{(i)\top}+\psi^2I_p$.
\end{enumerate}
We assume $\phi,\psi>0$ and $\sigma\ge0$. All expectations
below keep the partition and centroids fixed.

\begin{lemma}[Community levels and fluctuations]
\label{lem:lcm_moments}
For $j\ne k$,
\[
    \E_{\Sigma}\Sigma_{jk}=
    \begin{cases}
        \mu_W:=\phi^2d, & z_j=z_k,\\
        \mu_B:=0,      & z_j\ne z_k,
    \end{cases}
    \qquad
    \operatorname{Var}_{\Sigma}(\Sigma_{jk})
    =s^2:=2\sigma^2\phi^2d+\sigma^4d.
\]
On the diagonal,
\[
    m_D:=\E_{\Sigma}\Sigma_{jj}=d(\phi^2+\sigma^2)+\psi^2,
    \qquad
    \operatorname{Var}_{\Sigma}(\Sigma_{jj})=2s^2.
\]
For any $u>0$, with probability at least
$1-\{\binom{p}{2}+2p\}s^2/u^2$, simultaneously
\[
    \max_{j<k}|\Sigma_{jk}-\E_{\Sigma}\Sigma_{jk}|\le u,
    \qquad
    \max_j\Sigma_{jj}\le m_D+u.
\]
On this event, the measurement-noise factors satisfy
\[
    v_{jk}:=\Sigma_{jj}\Sigma_{kk}+\Sigma_{jk}^2
    \le 2(m_D+u)^2.
\]
\end{lemma}

\begin{proof}
Expanding $f_j^\top f_k$ and $\|f_j\|^2$ gives the moments
by independence and Gaussianity. Chebyshev's inequality
and union bound give the probability bound. Finally,
positive semidefiniteness gives
$\Sigma_{jk}^2\le\Sigma_{jj}\Sigma_{kk}$.
\end{proof}

At $\sigma=0$, every subject has the same connectome.
Increasing $\sigma$ introduces subject-level variation and
increases the spread $s$ relative to the fixed community
separation $\phi^2d$.

\subsubsection{Graph construction and threshold regimes}\label{app:lcm_threshold}
All four graphs are entrywise thresholdings $\mathbbm 1\{\cdot>\delta\}$ of a matrix ($\Sigma$, $\widehat\Sigma^\perp$, $\widehat\Sigma$ or $\Sigma_0$), so their structure is set by where $\delta$ sits relative to the between- and within-community levels $\mu_B$ and $\mu_W$ of $\Sigma$. We distinguish:
\begin{enumerate}[leftmargin=1.5em, itemsep=0pt, topsep=0pt]
    \item \textbf{Disconnected graph, $\delta=+\infty$.} $A=0$; after adding self-loops $R=I_p$, and the smoothed map is the raw map.
    \item \textbf{Complete graph, $\delta=-\infty$.} $\widetilde A=\mathbf 1\mathbf 1^\top$ and $R=P_{\mathbf 1}:=\frac1p\mathbf 1\mathbf 1^\top$, so $P_{\mathbf 1}\widehat\Sigma=\mathbf 1\widehat s^\top$ with $\widehat s:=\frac1p\widehat\Sigma\mathbf 1$: the $p^2$ entries collapse to the $p$ ROI averages.
    \item \textbf{Intermediate threshold in the regime, $\mu_B<\delta<\mu_W$,} with margin $\gamma_\delta:=\min\{\mu_W-\delta,\delta-\mu_B\}>0$. The oracle graph selects mostly within-community neighbors and
    \[
        R(A^{\text{oracle}})=P_C+\Delta R_\delta,
        \qquad
        (P_C)_{j\ell}=\frac1q\mathbbm 1\{\ell\in C(j)\},
    \]
    where $P_C$ depends only on the community partition and $\Delta R_\delta$ collects the subject-specific deviations of the oracle neighborhoods.
    \item \textbf{Finite threshold outside the regime.} If $\delta<\mu_B$, most between-community pairs are also connected: the graph is dense and $R$ approaches $P_{\mathbf 1}$. If $\delta>\mu_W$, most within-community pairs are dropped: the graph is sparse and $R$ approaches $I_p$, the surviving edges being the largest entries of $\Sigma^{(i)}$ rather than the partition. In neither case does $R$ split into a community and a subject-specific component.
\end{enumerate}
We focus on case 3 when interpreting the results through community structure. The oracle-smoothing bounds themselves hold for any fixed threshold.

\subsubsection{Predictive signal}\label{app:lcm_alignment}
Let $B^\star\in\mathbb R^{p\times p}$ be the matrix form of the regression coefficient, $\beta^\star=\operatorname{vec}(B^\star)$, so that $h(\Sigma)=\langle B^\star,\Sigma\rangle$. We say that $B^\star$ is community-smooth if rows from the same community have similar predictive profiles,
\[
    \max_j\max_{\ell\in C(j)}\sum_{k=1}^p\big(B^\star_{jk}-B^\star_{\ell k}\big)^2\le\kappa_\beta^2\quad\text{for some }\kappa_\beta\ge0;
\]
the ideal case $\kappa_\beta=0$ has $B^\star_{jk}$ depending only on $(z_j,z_k)$. We study two families of targets.
\begin{itemize}[leftmargin=1.5em, itemsep=0pt, topsep=0pt]
    \item \textbf{Targets constant over communities:} the within-community strength of a fixed ROI $a$, with $e_a$ the $a$-th standard basis vector and $\mathbf 1_{C(a)}$ the indicator of the community containing $a$,
    \[
        B^\star_{a,C}:=\tfrac12\big(e_a\mathbf 1_{C(a)}^\top+\mathbf 1_{C(a)}e_a^\top\big),\qquad h_{a,C}(\Sigma)=e_a^\top\Sigma\mathbf 1_{C(a)};
    \]
    and the total within-community connectivity $h_C(\Sigma)=|C|^{-2}\mathbf 1_C^\top\Sigma\mathbf 1_C$.
    \item \textbf{Targets supported on single edges or on within-community differences:} the single edge $h_{ab}(\Sigma)=\Sigma_{ab}$, and the within-community contrast
    \[
        B^\star_{\rm contrast}:=\tfrac12\{(e_a-e_b)v^\top+v(e_a-e_b)^\top\},\qquad v=\mathbf 1_{C(a)} .
    \]
\end{itemize}

\subsection{Results under the latent community model}
\subsubsection{Case 1: oracle graph}
\label{app:case-oracle}
The oracle graph is constructed from the true connectome. Specifically,
\[ A^{\text{oracle}}=\ind\{\Sigma>\delta\}, \qquad R(A^{\text{oracle}})=\tildD^{-1}(A^{\text{oracle}}+I_p). \]
Proposition~\ref{prop:eiv-message-passing}(a)
applies directly to this type of graph construction.

\medskip

\noindent\textbf{Proposition~\ref{prop:oracle_latent_community}
(Oracle smoothing).}
Under the EIV assumptions of Remark~\ref{thm:raw1}, fix any threshold $\delta$; the oracle operator $R(A^{\text{oracle}})$ is a function of $\Sigma$.
Assume all displayed moments are finite and define the squared bias and the variance reduction:
\begin{align*}
    \mathcal B_{A^{\text{oracle}}}(\delta)
    &:=
    \mathbb E_\Sigma
    \left[\langle B^\star,R(A^{\text{oracle}})\Sigma-\Sigma\rangle^2\right],\\
    \Delta\mathcal V_{A^{\text{oracle}}}(\delta)
    &:=
    T\,\mathbb E_\Sigma\left[
        \operatorname{Var}_\eta
        \{\langle B^\star,H\rangle\mid\Sigma\}
        -
        \operatorname{Var}_\eta
        \{\langle B^\star,R(A^{\text{oracle}}) H\rangle\mid\Sigma\}
    \right].
\end{align*}

\emph{Sufficient condition.}
The optimized population MSEs satisfy, for every $T>0$,
\[
    \operatorname{MSE}_{\mathrm{oracle}}(T,\delta)
    -\operatorname{MSE}_{\mathrm{raw}}(T)
    \le
    \mathcal B_{A^{\text{oracle}}}(\delta)
    -\frac{\Delta\mathcal V_{A^{\text{oracle}}}(\delta)}{T}
    +r_{\mathrm{raw}}(T),
\]
where
\[
    r_{\mathrm{raw}}(T):=
    T^{-2}\beta^{\star\top}
    \Gamma(S+T^{-1}\Gamma)^\dagger\Gamma\beta^\star
    \ge0.
\]
If the support of $\Sigma$ spans the symmetric matrices (Remark~\ref{thm:raw1}), as in the latent community model with $\sigma>0$, then $r_{\mathrm{raw}}(T)=O(T^{-2})$.
Thus oracle smoothing improves whenever
$\mathcal B_{A^{\text{oracle}}}(\delta)+r_{\mathrm{raw}}(T)
<\Delta\mathcal V_{A^{\text{oracle}}}(\delta)/T$.

\emph{Necessary condition.}
Let
\[
    b_\delta:=
    \min_{\gamma\in\mathbb R^{p^2}}
    \mathbb E_\Sigma\left[
        \left\{
        h(\Sigma)-\gamma^\top\operatorname{vec}(R(A^{\text{oracle}})\Sigma)
        \right\}^2
    \right]
\]
be the minimum noiseless prediction error, with $\gamma$ shared
across subjects. If $b_\delta>0$, then
\[
    \operatorname{MSE}_{\mathrm{oracle}}(T,\delta)
    <\operatorname{MSE}_{\mathrm{raw}}(T)
    \quad\Longrightarrow\quad
    T<\frac{\beta^{\star\top}\Gamma\beta^\star}{b_\delta}.
\]

\begin{proof}
Write $R=R(A^{\text{oracle}})$ and
$\widetilde x_{A^{\text{oracle}}}=\operatorname{vec}(R\widehat\Sigma)
=\operatorname{vec}(R\Sigma)+\operatorname{vec}(RH)$.
Since $R$ is a function of $\Sigma$ and $\mathbb E_\eta[\eta\mid\Sigma]=0$,
$\mathbb E_\eta[\operatorname{vec}(RH)\mid\Sigma]=0$, and $\varepsilon$
is independent of $(\Sigma,H)$, so for any $\gamma\in\mathbb R^{p^2}$
the cross terms vanish and
\begin{equation}\label{eq:oracle-dec}
    \mathbb E_{\Sigma,\eta,\varepsilon}\big[(Y-\gamma^\top\widetilde x_{A^{\text{oracle}}})^2\big]
    =\mathbb E_\Sigma\big[\{h(\Sigma)-\gamma^\top\operatorname{vec}(R\Sigma)\}^2\big]
    +\mathbb E_\Sigma\big[\operatorname{Var}_\eta(\gamma^\top\operatorname{vec}(RH)\mid\Sigma)\big]
    +\sigma_\varepsilon^2 .
\end{equation}
Evaluating at $\gamma=\beta^\star$, rather than at the fitted
coefficient, the first term is $\mathcal B_{A^{\text{oracle}}}$ and the
second is
$T^{-1}(\beta^{\star\top}\Gamma\beta^\star-\Delta\mathcal V_{A^{\text{oracle}}})$,
by definition of $\Delta\mathcal V_{A^{\text{oracle}}}$ and
$T\,\mathbb E_\Sigma[\operatorname{Var}_\eta(\langle B^\star,H\rangle\mid\Sigma)]
=\beta^{\star\top}\Gamma\beta^\star$. Hence
\[
    \operatorname{MSE}_{\mathrm{oracle}}-\sigma_\varepsilon^2
    \le
    \mathcal B_{A^{\text{oracle}}}
    +T^{-1}\big(
        \beta^{\star\top}\Gamma\beta^\star
        -\Delta\mathcal V_{A^{\text{oracle}}}
    \big).
\]
For raw features, Remark~\ref{thm:raw1}, with $M=S+T^{-1}\Gamma$, gives
$\operatorname{MSE}_{\mathrm{raw}}-\sigma_\varepsilon^2
=\beta^{\star\top}(S-SM^\dagger S)\beta^\star$.
Substituting $S=M-T^{-1}\Gamma$, valid since $M\succeq T^{-1}\Gamma$, gives
\[
    S-SM^\dagger S=T^{-1}\Gamma-T^{-2}\Gamma M^\dagger\Gamma,
    \qquad\text{hence}\qquad
    \operatorname{MSE}_{\mathrm{raw}}-\sigma_\varepsilon^2
    =T^{-1}\beta^{\star\top}\Gamma\beta^\star-r_{\mathrm{raw}}(T),
\]
with $r_{\mathrm{raw}}(T)\ge0$ because $M^\dagger\succeq0$.
Subtracting proves the sufficient comparison.

By \eqref{eq:oracle-dec},
$\operatorname{MSE}_{\mathrm{oracle}}-\sigma_\varepsilon^2\ge b_\delta$, while
$\operatorname{MSE}_{\mathrm{raw}}-\sigma_\varepsilon^2\le T^{-1}\beta^{\star\top}\Gamma\beta^\star$.
Strict improvement therefore gives
$b_\delta<T^{-1}\beta^{\star\top}\Gamma\beta^\star$, the necessary bound on $T$.

Finally, under the spanning condition $S$ is invertible on the
symmetric subspace, which contains $\operatorname{range}(\Gamma)$
(Definition~\ref{def:commutation}). There $M\succeq S$, so
$M^\dagger\preceq S^\dagger$ and
\[
    r_{\mathrm{raw}}(T)
    \le T^{-2}\beta^{\star\top}\Gamma S^\dagger\Gamma\beta^\star
    =O(T^{-2}).
\]
In the latent community model with $\sigma>0$, taking $F=ue_1^\top$ shows that the support
of $\Sigma$ contains $uu^\top+\psi^2 I_p$ for every $u\in\mathbb R^p$,
and these matrices span the symmetric matrices.
\end{proof}

The sufficient condition above can now be interpreted through
the three ingredients that determine its bias and variance terms.

\textbf{Predictive signal, $B^\star$.}
Since the rows of $R(A^{\text{oracle}})$ sum to one,
\[
    \langle B^\star,R(A^{\text{oracle}})\Sigma-\Sigma\rangle
    =
    \sum_{j,k}B^\star_{jk}\sum_{\ell}
    R(A^{\text{oracle}})_{j\ell}\big(\Sigma_{\ell k}-\Sigma_{jk}\big),
\]
so the bias is small when the ROIs averaged with $j$ have rows
close to $\Sigma_{j\cdot}$ on the support of $B^\star$: community-level targets depend on sums of entries, whereas
edge targets depend on specific entries.

\textbf{Threshold, $\delta$.}
For a given target, the threshold determines which ROIs enter
this average. Lowering $\delta$ adds neighbors while raising it removes them.
These changes affect both bias and noise propagation, so more
averaging is not automatically better.

\emph{Disconnected graph.}
For $\delta=+\infty$, $R(A^{\text{oracle}})=I_p$
(Appendix~\ref{app:lcm_threshold}): the smoothed and raw features
coincide, so
$\operatorname{MSE}_{\mathrm{oracle}}(T,+\infty)
=\operatorname{MSE}_{\mathrm{raw}}(T)$ for every $T$, and
$\mathcal B_{A^{\text{oracle}}}=\Delta\mathcal V_{A^{\text{oracle}}}=0$.

\emph{Complete graph.}
For $\delta=-\infty$, $R(A^{\text{oracle}})=P_{\mathbf1}$
(Appendix~\ref{app:lcm_threshold}) is the same for every subject
and $P_{\mathbf1}\widehat\Sigma=\mathbf1\widehat s^\top$,
so the smoothed class is the set of linear functions of the
degree vector $\widehat s=\widehat\Sigma\mathbf1/p$.
By Proposition~\ref{prop:no_linear_gain},
$\operatorname{MSE}_{J}(T)\ge\operatorname{MSE}_{\mathrm{raw}}(T)$
for every $T$, with $J=\mathbf1\mathbf1^\top$.

\emph{Intermediate threshold.}
For $\mu_B<\delta<\mu_W$, the population operator is $P_C$.
Writing $R(A^{\text{oracle}})=P_C+\Delta R_\delta$ gives
\[
    \mathcal B_{A^{\text{oracle}}}(\delta)
    =
    \mathbb E_\Sigma\!\left[
        \left\{
        \langle B^\star,(P_C-I_p)\Sigma\rangle
        +
        \langle B^\star,\Delta R_\delta\Sigma\rangle
        \right\}^{2}
    \right].
\]
The first term concerns fixed community averaging; the second
captures the effect of subject-specific neighborhoods.
For exact community smoothness, $P_CB^\star=B^\star$, the first
term vanishes, but the second need not.
Since fixed averaging cannot improve optimized prediction
(Proposition~\ref{prop:no_linear_gain}), any oracle gain requires
subject-specific variation in the operator and not merely an
aligned target.

\textbf{Community structure, $\sigma$.}
At a given threshold, changing $\sigma$ changes the connectomes
being thresholded. It controls deviations from the shared
community loadings, affecting both the selected neighborhoods
and the measurement-noise factors.
For intermediate thresholds, concentration around the community
levels relative to the threshold margins makes the oracle close
to $P_C$. Thus clear communities and an aligned target alone
do not guarantee improvement: the prediction effect depends
on how $\sigma$, $\delta$, and $B^\star$ interact.

Together, these ingredients determine whether the noise reduction outweighs the squared bias and the raw-regression correction.
The two targets below illustrate this interaction.

\textbf{Concrete examples.} Consider $h_{a,v}(\Sigma)=e_a^\top\Sigma v$, taking $v=\mathbf 1_{C(a)}$ for the within-community strength or $v=e_b$ for a single edge. On the oracle-smoothed features, consider the candidate predictor
\[
    \widehat h_\delta
    =e_a^\top R(A^{\text{oracle}})\widehat\Sigma v
    =w_{a,\delta}^\top\widehat\Sigma v,
    \qquad
    w_{a,\delta}:=R(A^{\text{oracle}})^\top e_a .
\]
This is a linear predictor on $\operatorname{vec}(R(A^{\text{oracle}})\widehat\Sigma)$ with fixed coefficient $\operatorname{vec}(e_av^\top)$; it need not be the optimal fitted coefficient. Its conditional bias and measurement-noise variance factor are
\[
\begin{aligned}
    b_\delta(a,v;\Sigma)
    &=(w_{a,\delta}-e_a)^\top\Sigma v,\\
    V_\delta(a,v;\Sigma)
    &=(w_{a,\delta}^\top\Sigma w_{a,\delta})(v^\top\Sigma v)
     +(w_{a,\delta}^\top\Sigma v)^2 .
\end{aligned}
\]
Write $V_0$ for the same expression with $w_{a,\delta}=e_a$. Conditional centering and the Wishart moments give
\[
    \E_{\eta,\varepsilon}\big[(y-\widehat h_\delta)^2\mid\Sigma\big]
    =b_\delta(a,v;\Sigma)^2+\frac{V_\delta(a,v;\Sigma)}{T}
    +\sigma_\varepsilon^2 .
\]
Optimizing the oracle regression can only decrease this risk. Since the optimized raw risk is $\sigma_\varepsilon^2+T^{-1}\E_\Sigma[V_0]-r_{\mathrm{raw}}(T)$, the argument of Proposition~\ref{prop:oracle_latent_community} guarantees improvement whenever
\[
    \E_\Sigma[b_\delta(a,v;\Sigma)^2]+r_{\mathrm{raw}}(T)
    <
    \frac{\E_\Sigma[V_0(a,v;\Sigma)-V_\delta(a,v;\Sigma)]}{T},
\]
where $r_{\mathrm{raw}}(T)$ is evaluated for the chosen target.

\subsubsection{Case 2: independent-copy graph and graph-estimation error}
\label{app:case-independent-copy}

The independent graph is built from a second estimate
\(\widehat\Sigma^\perp=\Sigma+H^\perp\) with \(H^\perp\indep H\mid\Sigma\) so that
\(A^\perp=\ind\{\widehat\Sigma^\perp>\delta\}\). By Lemma~\ref{lem:marginal} it has the same
conditional law as \(\widehat A\) given \(\Sigma\) but is independent of the feature noise, so
Proposition~\ref{prop:eiv-message-passing}(a) applies with \(A=A^\perp\). The graph-estimation contribution is
\(\mse_{\mathrm{ind}}(T)-\mse_{\mathrm{oracle}}(T)\): all differences relative to the oracle come from
smoothing over the wrong neighborhoods and not from coupling with \(H\).

\medskip
\noindent\textbf{Proposition~\ref{prop:independent_graph_window}
(Graph-estimation error for the independent graph).}
Under the assumptions of Proposition~\ref{prop:oracle_latent_community}, let $A^\perp$ be built from a conditionally independent copy $H^\perp$ of $H$ given $\Sigma$, and assume $\mathbb E_\Sigma\|\Sigma\|_F^4<\infty$.
Define the edge-error probability and the graph-estimation error:
\[
    \rho(T,\delta):=\mathbb P(A^\perp\ne A^{\text{oracle}}),
    \qquad
    \mathcal E(T,\delta):=C(1+T^{-1})\sqrt{\rho(T,\delta)},
\]
where $C$ depends only on $p$, $\|B^\star\|_F$ and $\mathbb E_\Sigma\|\Sigma\|_F^4$, not on $T$ or $\delta$. Then, for every $T>0$,
\[
    \operatorname{MSE}_{\mathrm{ind}}(T,\delta)
    -\operatorname{MSE}_{\mathrm{raw}}(T)
    \le
    \mathcal B_{A^{\text{oracle}}}(\delta)
    -\frac{\Delta\mathcal V_{A^{\text{oracle}}}(\delta)}{T}
    +\mathcal E(T,\delta)+r_{\mathrm{raw}}(T).
\]
Thus the independent graph improves whenever
$\mathcal B_{A^{\text{oracle}}}(\delta)+\mathcal E(T,\delta)+r_{\mathrm{raw}}(T)
<\Delta\mathcal V_{A^{\text{oracle}}}(\delta)/T$.

\begin{proof}
As in the proof of Proposition~\ref{prop:oracle_latent_community}, we evaluate both graphs at the true coefficient $\beta^\star$ instead of the fitted one. Given $\Sigma$ and the graph $A$, the resulting error, excluding the response noise, is
\[
    L_A=\langle B^\star,R(A)\Sigma-\Sigma\rangle^2
    +\operatorname{Var}_\eta\{\langle B^\star,R(A)H\rangle\mid\Sigma,A\}.
\]
For the oracle, its expectation is $\mathcal B_{A^{\text{oracle}}}+T^{-1}(\beta^{\star\top}\Gamma\beta^\star-\Delta\mathcal V_{A^{\text{oracle}}})$. Hence
\[
    \operatorname{MSE}_{\mathrm{ind}}-\sigma_\varepsilon^2
    \le\mathbb E_{\Sigma,H^\perp}[L_{A^\perp}]
    =\mathbb E_\Sigma[L_{A^{\text{oracle}}}]
    +\mathbb E_{\Sigma,H^\perp}[L_{A^\perp}-L_{A^{\text{oracle}}}].
\]
The last term is the graph-estimation error. It is zero when $A^\perp=A^{\text{oracle}}$. Since $\|R(A)\|_{\mathrm{op}}\le\sqrt p$ and $\|\Psi\|_{\mathrm{op}}\le2\|\Sigma\|_F^2$, each loss is at most $C_0(1+T^{-1})\|\Sigma\|_F^2$ with $C_0=4p\|B^\star\|_F^2$, and Cauchy--Schwarz bounds the last term by $\mathcal E$ with $C=C_0(\mathbb E_\Sigma\|\Sigma\|_F^4)^{1/2}$. Subtracting $\operatorname{MSE}_{\mathrm{raw}}-\sigma_\varepsilon^2=T^{-1}\beta^{\star\top}\Gamma\beta^\star-r_{\mathrm{raw}}(T)$ (proof of Proposition~\ref{prop:oracle_latent_community}) gives the bound.
\end{proof}

Proposition~\ref{prop:independent_graph_window} asks whether
graph estimation is accurate enough to retain the improvement
certified for the oracle.

\textbf{Predictive signal, $B^\star$.}
The target determines the oracle bias-variance tradeoff and
the predictive consequences of changing its neighborhoods,
but not the edge-error probabilities.
In the preceding examples, neighborhood errors change averages
of community strengths or individual connectivity entries.
The same graph-estimation accuracy therefore need not give
the same prediction benefit for different targets.

\textbf{Threshold, $\delta$.}
The threshold determines both the oracle neighborhoods and
how reliably they can be estimated.
For the following calculation, adopt the Gaussian surrogate
$\widehat\Sigma^\perp_{jk}\mid\Sigma
\sim\mathcal N(\Sigma_{jk},v_{jk}/T)$, where
$v_{jk}=\Sigma_{jj}\Sigma_{kk}+\Sigma_{jk}^2>0$.
Standardizing the threshold-crossing event gives
\[
    \pi_{jk}:=
    \mathbb P(A^\perp_{jk}\ne A^{\text{oracle}}_{jk}\mid\Sigma)
    =
    \Phi\!\left(
        -\frac{|\Sigma_{jk}-\delta|\sqrt T}{\sqrt{v_{jk}}}
    \right),
\]
where $\Phi$ is the standard normal CDF.
Reliability therefore depends on actual threshold margins
relative to measurement noise, not on graph density alone.

\emph{Disconnected and complete graphs.}
At $\delta=+\infty$ both operators equal $I_p$; at
$\delta=-\infty$ both equal $P_{\mathbf1}$.
In either case, $\rho=\mathcal E=0$.
These graphs are perfectly estimated, but cannot improve
optimized prediction over raw features.

\emph{Intermediate thresholds.}
For $\mu_B<\delta<\mu_W$, the oracle decomposition gives
\[
    R(A^\perp)
    =
    P_C+\Delta R_\delta
    +\big\{R(A^\perp)-R(A^{\text{oracle}})\big\}.
\]
The last term is the graph-estimation change; $\Delta R_\delta$
is genuine subject-specific variation, not estimation error.
The aim is therefore to recover the oracle neighborhoods,
not merely the population partition.

Set $\gamma_\delta=\min\{\mu_W-\delta,\delta-\mu_B\}$.
Suppose an event $G$, with $\mathbb P(G^c)\le\epsilon_0$,
ensures that all off-diagonal entries lie within
$\gamma_\delta/2$ of their community levels and
$v_{jk}\le v_{\max}:=\max\{v_W,v_B\}$, with deterministic
bounds independent of $T$.
On $G$, the actual margins are at least $\gamma_\delta/2$,
so the Gaussian tail bound gives
\[
    \pi_{jk}\le e^{-T/(8T_{\mathrm{graph}}(\delta))},
    \qquad
    T_{\mathrm{graph}}(\delta):=
    \frac{v_{\max}}{\gamma_\delta^2}.
\]
A union bound over edges, including the exceptional event, yields
\[
    \rho(T,\delta)
    \le
    \min\left\{
        1,\,
        \epsilon_0+\binom p2
        e^{-T/(8T_{\mathrm{graph}}(\delta))}
    \right\}.
\]

\textbf{Community structure, $\sigma$.}
With fixed centroids, changing $\sigma$ leaves $\mu_W,\mu_B$
and $\gamma_\delta$ unchanged, but changes concentration around
those levels and the measurement-noise scales.
Lemma~\ref{lem:lcm_moments} makes this dependence explicit:
the preceding event can be obtained with
\[
    \epsilon_0=
    \min\left\{1,\,
        \frac{4\{\binom p2+2p\}s^2}{\gamma_\delta^2}
    \right\},
    \qquad
    v_W=v_B=2(m_D+\gamma_\delta/2)^2.
\]
The community-margin guarantee is useful only when its
exceptional probability is small; an intermediate threshold
alone does not ensure reliable graph estimation.

Together, these ingredients determine whether the
graph-estimation bound fits within the oracle's sufficient
improvement margin.
Increasing $T$ improves edge estimation under the surrogate,
but also reduces the raw-feature noise available to denoise.
An intermediate favorable range may therefore arise,
but is not guaranteed. Failure of the sufficient condition
remains inconclusive.

\subsubsection{Case 3: same-sample graph and double dipping}
\label{app:case-estimated-graph}

The standard pipeline thresholds the same empirical covariance that serves as node
features, \(\widehat A=\ind\{\widehat\Sigma>\delta\}\), so \(\cR_{\widehat A}\) depends on
\(H\).
By Lemma~\ref{lem:marginal} the independent graph has the same conditional law given
\(\Sigma\) but none of the couplings, so
\(\mathcal D_{\rm dd}(T,\delta)=\mse_{\mathrm{same}}(T)-\mse_{\mathrm{ind}}(T)\) isolates the
double-dipping contribution.

\medskip

\noindent\textbf{Proposition~\ref{prop:double_dipping_window}
(Double-dipping contribution for the same-sample graph).}
Under the assumptions of Proposition~\ref{prop:independent_graph_window}, define the double-dipping contribution $\mathcal D_{\rm dd}(T,\delta):=\operatorname{MSE}_{\mathrm{same}}(T,\delta)-\operatorname{MSE}_{\mathrm{ind}}(T,\delta)$. Then, for every $T>0$,
\[
    \operatorname{MSE}_{\mathrm{same}}(T,\delta)
    -\operatorname{MSE}_{\mathrm{raw}}(T)
    \le
    \mathcal B_{A^{\text{oracle}}}(\delta)
    -\frac{\Delta\mathcal V_{A^{\text{oracle}}}(\delta)}{T}
    +\mathcal E(T,\delta)+\mathcal D_{\rm dd}(T,\delta)+r_{\mathrm{raw}}(T).
\]
Thus the same-sample graph improves whenever
$\mathcal B_{A^{\text{oracle}}}(\delta)+\mathcal E(T,\delta)+\mathcal D_{\rm dd}(T,\delta)+r_{\mathrm{raw}}(T)
<\Delta\mathcal V_{A^{\text{oracle}}}(\delta)/T$.
The contribution $\mathcal D_{\rm dd}$ is signed: negative values favor the same-sample graph.

\begin{proof}
Since $\operatorname{MSE}_{\mathrm{same}}=\operatorname{MSE}_{\mathrm{ind}}+\mathcal D_{\rm dd}$ by definition, the bound follows from Proposition~\ref{prop:independent_graph_window}.
\end{proof}

\medskip
\noindent\textbf{Size of the double-dipping contribution.}
Suppose in addition that $\mathbb E_{\Sigma,\eta}\|H\|_F^4\le C_H^4T^{-2}$ and that the positive eigenvalues of $M_{\widehat A}$ and $M_{A^\perp}$ are at least $\lambda>0$ for all $T\ge1$. Then, for $T\ge1$,
\[
    |\mathcal D_{\rm dd}(T,\delta)|
    \le C_{\rm dd}\sqrt{\frac{\rho(T,\delta)}{T}},
\]
where $C_{\rm dd}$ depends only on $p$, $\|B^\star\|_F$, $C_H$, $\lambda$ and $\mathbb E_\Sigma\|\Sigma\|_F^4$, not on $T$.

\begin{proof}
Let $F_A(\gamma):=\mathbb E_\Sigma[h^2]-2\gamma^\top c_A+\gamma^\top M_A\gamma$, with $c_A$ and $M_A$ as in Proposition~\ref{prop:eiv-message-passing} and $M_{A^\perp}=S_{A^\perp}+T^{-1}\Gamma_{A^\perp}$, so that $\operatorname{MSE}_A=\sigma_\varepsilon^2+\min_\gamma F_A(\gamma)$. Evaluating each risk at the other graph's minimizer gives
\[
    q(\beta^*_{\widehat A})\le\mathcal D_{\rm dd}\le q(\beta^*_{A^\perp}),
    \qquad
    q(\gamma):=F_{\widehat A}(\gamma)-F_{A^\perp}(\gamma)=\gamma^\top\Delta_M\gamma-2\gamma^\top\Delta_c,
\]
with $\Delta_M:=M_{\widehat A}-M_{A^\perp}$ and $\Delta_c:=c_{\widehat A}-c_{A^\perp}$. The graphs $\widehat A$ and $A^\perp$ have the same law given $\Sigma$ (Lemma~\ref{lem:marginal}), so every term without $\eta$ cancels and only the noise reuse remains:
\[
    \Delta_c=\mathbb E_{\Sigma,\eta}[(\mathcal R_{\widehat A}-\mathcal R_{A^{\text{oracle}}})\eta h],
    \qquad
    \Delta_M=\mathbb E_{\Sigma,\eta}[\mathcal R_{\widehat A}Q_T\mathcal R_{\widehat A}^\top-\mathcal R_{A^{\text{oracle}}}Q_T\mathcal R_{A^{\text{oracle}}}^\top],
\]
where $Q_T:=\widehat x\widehat x^\top-xx^\top-T^{-1}\Psi$. The oracle terms have mean zero because $\mathbb E_\eta[\eta\mid\Sigma]=0$ and $\mathbb E_\eta[Q_T\mid\Sigma]=0$. Both integrands vanish unless $\widehat A\ne A^{\text{oracle}}$, an event of probability $\rho$. Cauchy--Schwarz with $\|\mathcal R_A\|_{\mathrm{op}}\le\sqrt p$ and the fourth moments gives $\|\Delta_c\|+\|\Delta_M\|_{\mathrm{op}}\le C_1\sqrt{\rho/T}$. Finally, $\lambda\|\beta^*_A\|^2\le\beta^{*\top}_AM_A\beta^*_A\le\mathbb E_\Sigma[h^2]$ for both graphs, hence
\[
    |q(\beta^*_A)|
    \le
    \frac{\mathbb E_\Sigma[h^2]}{\lambda}\|\Delta_M\|_{\mathrm{op}}
    +2\sqrt{\frac{\mathbb E_\Sigma[h^2]}{\lambda}}\|\Delta_c\|,
\]
which gives the bound.
\end{proof}

Since $\widehat A$ and $A^\perp$ have the same law given $\Sigma$, $\mathcal D_{\rm dd}$ isolates noise reuse, not a difference in graph quality. For $\mu_B<\delta<\mu_W$, both graphs estimate $R(A^{\text{oracle}})=P_C+\Delta R_\delta$. The subject-specific part $\Delta R_\delta$ is not double dipping. What differs is that $R(\widehat A)-R(A^{\text{oracle}})$ depends on the feature noise $H$, while $R(A^\perp)-R(A^{\text{oracle}})$ does not. The bound shows that double dipping lives on edge flips: it is zero at $\delta=\pm\infty$, where both operators equal $I_p$ or $P_{\mathbf1}$, and small whenever $\rho$ is small. Its sign depends on $B^\star$, and community smoothness alone fixes neither its sign nor its size. When the bound is small, the two optimized errors are close: removing noise reuse changes little, and the shared graph-estimation error remains.

\section{Additional simulation results}\label{app:simulations}
In this section we report additional simulation experiments that further validate our theoretical findings and extend them to settings beyond the scope of the main analysis.

We begin by showing that the same conclusions hold when the feature map is the sample correlation $\widehat C$ rather than the sample covariance $\widehat\Sigma$, the more common choice in fMRI pipelines. We then provide ablations along the latent structure $\sigma$, the threshold $\delta$, and the regression target $\beta$, isolating how each axis modulates the three effects identified in Section~\ref{sec:specialization}. Departing from the population regime ($n \to \infty$), we show that at small $n$ even smoothing with the correct graph topology may fail to improve over raw features. We further demonstrate that the same statistical framework applies to GCN \cite{kipf2017semisupervisedclassificationgraphconvolutional} without changing the qualitative ordering between feature maps. Finally, we extend the analysis from i.i.d.\ samples to autoregressive time series, which more faithfully reflect the temporal structure of real fMRI signals~\citep{Worsleyetal2002}; this departure preserves the framework and amounts to a shift along the $T$ axis driven by the reduction in effective sample size.

\subsection{Extension to correlation matrices: $C$} \label{sec:appendix-correlation}
\begin{figure}[htb]
    \centering
    \includegraphics[width=\linewidth]{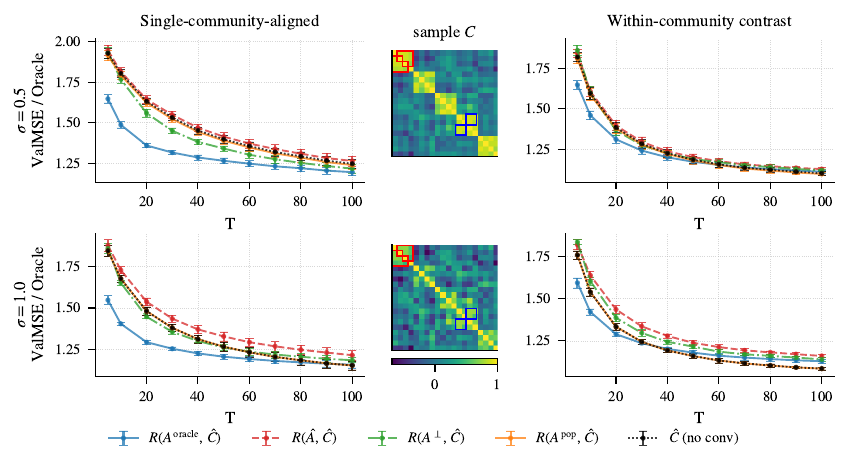}
    \vspace{-18pt}
    \caption{\textbf{Decoupling the effects for different values of $\beta$ and latent structures, with $C$ in place of $\Sigma$.} Center: true correlation $C$ for two graph regimes ($\sigma \in \{0.5, 1.0\}$); overlays mark the support of $\beta=\texttt{single\_community\_aligned}$ (red, full community block) and $\beta=\texttt{within\_community\_contrast}$ (blue, single in-community edge). Left/right: normalized ValMSE vs.\ $T$ for $\mathrm{SageMean}(A, \widehat C)$ under $A \in \{A^{\text{oracle}}, \widehat A, A^\perp, A^{\mathrm{pop}}\}$ and a no-conv baseline. Mean $\pm$ 1 SD over 5 seeds. $p=20$, $K=5$, $n=25{,}000$; thresholds $\delta_{\sigma=0.5}\approx 0.79$, $\delta_{\sigma=1.0}\approx 0.53$.}
    \label{fig:main-C}
\end{figure}

The theoretical analysis in Section~\ref{sec:stats_framework} is stated for the sample covariance $\widehat\Sigma$, but fMRI pipelines almost always operate on the sample correlation $\widehat C = D^{-1/2}\widehat\Sigma D^{-1/2}$. Figure~\ref{fig:main-C} reproduces the experiment of Figure~\ref{fig:main} with $\widehat C$ as the feature map and confirms that the ordering between feature maps is preserved.

\subsection{The role of the coefficient: $\beta$}
\begin{figure}
    \centering
    \includegraphics[width=\linewidth]{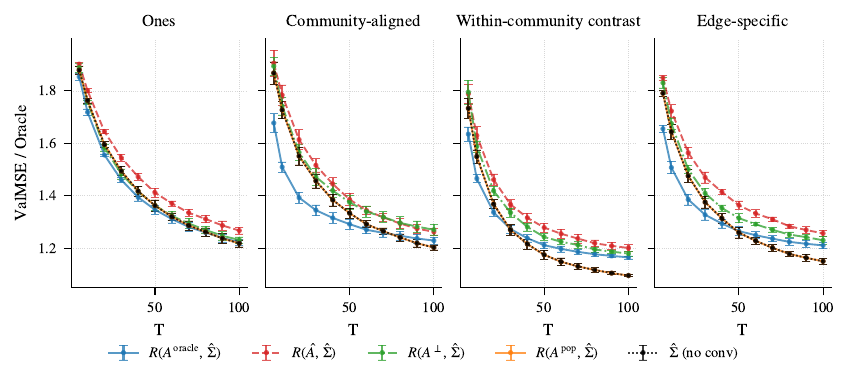}
    \vspace{-18pt}
    \caption{\textbf{Decoupling the effect of the regression target $\beta$.} Each panel fixes a different $\beta$ pattern (\texttt{ones}, \texttt{community\_aligned}, \texttt{within\_community\_contrast}, \texttt{edge\_specific}) and plots normalized ValMSE vs.\ $T$ for $\mathrm{SageMean}(A, \widehat\Sigma)$ under $A \in \{A^{\text{oracle}}, \widehat A, A^\perp, A^{\mathrm{pop}}\}$ and the no-conv baseline. Mean $\pm$ 1 SD over 5 seeds. $\sigma=1.0$, $p=20$, $K=5$, $n=25{,}000$, threshold $\delta\approx 10.20$.}
    \label{fig:beta-effect}
\end{figure}

Fixing the latent structure ($\sigma=1.0$) and a sparse threshold ($\delta\approx 10.20$), we vary the regression target $\beta$. Figure~\ref{fig:beta-effect} shows that smoothing helps when $\beta$ is community-smooth (\texttt{community\_aligned}): the within-community averaging operator induces little bias and reduces variance, so $A^{\text{oracle}}$ improves over raw features across all $T$. For the \texttt{within\_community\_contrast} and \texttt{edge\_specific} targets, smoothing averages together precisely the entries whose differences carry the predictive signal, and the bias dominates the variance reduction except at very small $T$.

\subsection{Different types of convolutions}
\label{sec:appendix-conv}
\begin{figure}
    \centering
    \includegraphics[width=\linewidth]{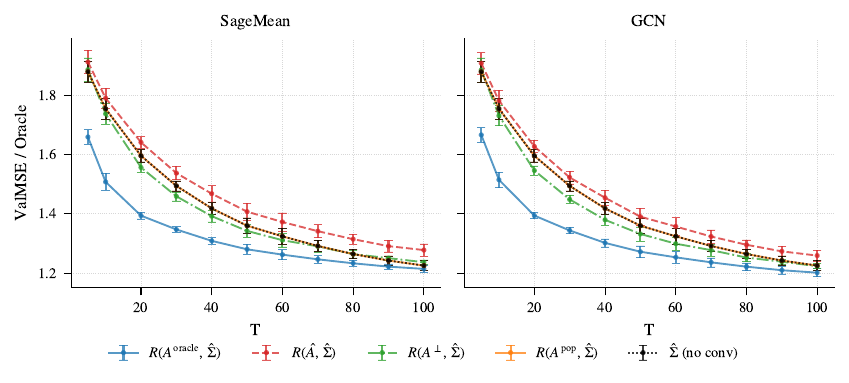}
    \vspace{-18pt}
    \caption{\textbf{Decoupling the effect of the convolution operator.} Each panel fixes a different graph convolution (\textsc{SageMean}, \textsc{GCN}) and plots normalized ValMSE vs.\ $T$ for $\mathrm{Conv}(A, \widehat\Sigma)$ under $A \in \{A^{\text{oracle}}, \widehat A, A^\perp, A^{\mathrm{pop}}\}$ and the no-conv baseline. Mean $\pm$ 1 SD over 5 seeds. $\beta=\texttt{single\_community\_aligned}$, $\sigma=1.0$, $p=20$, $K=5$, $n=25{,}000$, $k=1$ convolution layer, threshold $\delta\approx 10.20$.}
    \label{fig:conv-sweep}
\end{figure}
The analysis in Section~\ref{sec:specialization} uses the SageMean operator $R_{\rm SAGE} = \widetilde D^{-1}\widetilde A$~\citep{hamilton2018inductiverepresentationlearninglarge}. We consider its linearized GCN counterpart $R_{\rm GCN} = \widetilde D^{-1/2}\widetilde A\widetilde D^{-1/2}$~\citep{kipf2017semisupervisedclassificationgraphconvolutional}. Figure~\ref{fig:conv-sweep} shows that SageMean and GCN behave near-identically: with balanced communities and a moderate $\delta$, neighborhoods have roughly equal cardinality, so row- and symmetric-normalization coincide up to a constant. The qualitative ordering between $A^{\text{oracle}}$, $A^\perp$, $\widehat A$, and $A^{\mathrm{pop}}$ is preserved across both operators.

\subsection{Temporally correlated samples}
\begin{figure}
    \centering
    \includegraphics[width=\linewidth]{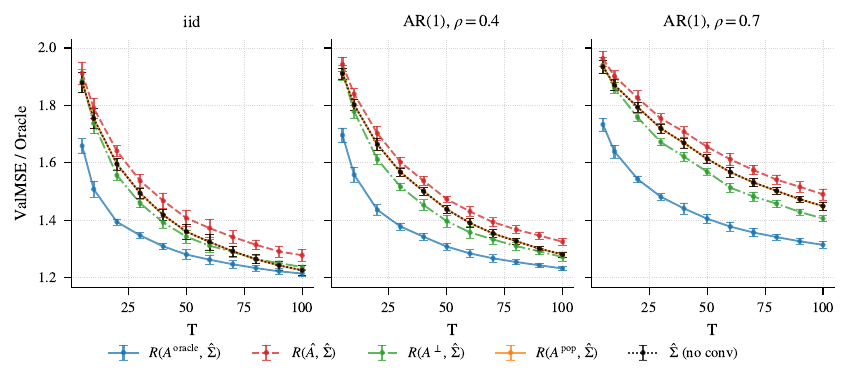}
    \vspace{-18pt}
    \caption{\textbf{Decoupling the effect of temporal correlation.} Each panel fixes a different temporal-noise regime for the time series used to estimate $\widehat\Sigma$, iid samples (left), AR(1) with $\rho=0.4$ (center), AR(1) with $\rho=0.7$ (right), and plots normalized ValMSE vs.\ $T$ for $\mathrm{SageMean}(A, \widehat\Sigma)$ under $A \in \{A^{\text{oracle}}, \widehat A, A^\perp, A^{\mathrm{pop}}\}$ and the no-conv baseline. Mean $\pm$ 1 SD over 5 seeds. $\beta=\texttt{single\_community\_aligned}$, $\sigma=1.0$, $p=20$, $K=5$, $n=25{,}000$, $k=1$ convolution layer, threshold $\delta\approx 10.20$.}
    \label{fig:sim-sweep}
\end{figure}

The analysis in Section~\ref{sec:specialization} assumes i.i.d.\ samples $Z_t \sim \mathcal{N}(0, \Sigma)$, but real fMRI time series are temporally autocorrelated~\citep{Worsleyetal2002}. We replace the i.i.d.\ samples with AR(1) processes at $\rho \in \{0.4, 0.7\}$. For an AR(1) process the effective sample size is $T_{\rm eff} \approx T(1-\rho^2)/(1+\rho^2)$, so $\widehat\Sigma$ is noisier at fixed $T$. Figure~\ref{fig:sim-sweep} confirms this: the curves for $\rho=0.4$ and $\rho=0.7$ are shifted to the right relative to the i.i.d.\ panel, while the relative ordering between $A^{\text{oracle}}$, $A^\perp$, $\widehat A$, and $A^{\mathrm{pop}}$ is preserved. Temporal correlation therefore rescales rather than reshapes the bias-variance tradeoff identified in the main analysis.

\subsection{Small sample size}
\begin{figure}
    \centering
    \includegraphics[width=\linewidth]{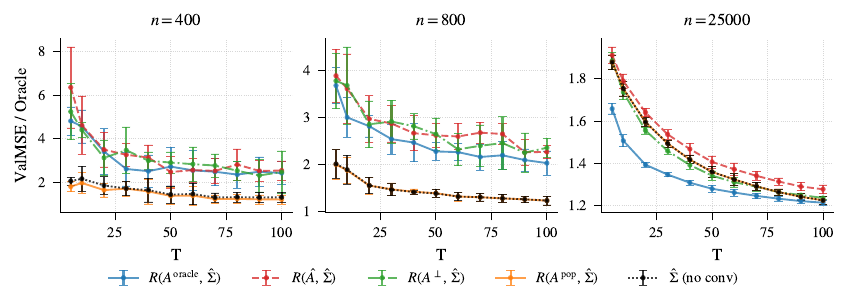}
    \vspace{-18pt}
    \caption{\textbf{Decoupling the effect of the sample size $n$.} Each panel fixes a different number of training samples $n \in \{400, 800, 25000\}$ and plots normalized ValMSE vs.\ $T$ for $\mathrm{SageMean}(A, \widehat\Sigma)$ under $A \in \{A^{\text{oracle}}, \widehat A, A^\perp, A^{\mathrm{pop}}\}$ and the no-conv baseline. Mean $\pm$ 1 SD over 5 seeds. $\beta=\texttt{single\_community\_aligned}$, $\sigma=1.0$, $p=20$, $K=5$, threshold $\delta\approx 10.20$.}
    \label{fig:n-sweep}
\end{figure}

Our analysis restricts to the population regime $n \to \infty$. Figure~\ref{fig:n-sweep} departs from this assumption and varies $n \in \{400, 800, 25{,}000\}$. At $n = 25{,}000$ the population predictions of the main analysis are recovered cleanly. At smaller $n$, finite-sample estimation variance becomes the dominant source of error and the ordering between feature maps is no longer well predicted by the EIV decomposition: smoothing with the correct topology no longer guarantees an improvement over raw features.

\section{Real data}\label{app:real_data}
\subsection{Datasets}
\subsubsection{ADHD dataset}\label{app:adhd_dataset}
The ADHD-200 dataset is a multi-site collection of resting-state functional magnetic resonance imaging data targeting attention deficit hyperactivity disorder identification and cognitive assessment \cite{ADHD200Consortium2012}. We utilized the preprocessed data provided by the Neuro Bureau ADHD-200 Preprocessed repository \cite{bellec2017neuro}.
Specifically, the imaging data were processed using the Athena pipeline. Regional mean time series were extracted using the Craddock 200 (CC200) atlas \cite{Craddock2012}, with 190 retained regions and scan lengths ranging from 55 to 257 time points across 1,198 scans from 768 subjects. Acquisition parameters vary by site, with repetition times (TR) ranging from $1.96$ to $2.5$ seconds \cite{bellec2017neuro}.
For our analysis, we focus on the following target variables:
\begin{itemize}
    \item \texttt{Diagnosis}: The primary clinical label used for classification, distinguishing between individuals with attention deficit hyperactivity disorder and typical controls.
    \item \texttt{Inattentive}: The inattention symptom score reported in the ADHD-200 phenotypic file, used for regression.
\end{itemize}

\subsubsection{ABIDE dataset}\label{app:abide_dataset}
The Autism Brain Imaging Data Exchange (ABIDE) aggregates resting-state functional magnetic resonance imaging data from 17 international sites to investigate Autism Spectrum Disorder (ASD) \cite{dimartino2014autism}. We utilized the preprocessed data provided by the Preprocessed Connectomes Project (PCP) \cite{craddock2013neuro}.
Specifically, imaging data were processed using the Configurable Pipeline for the Analysis of Connectomes (C-PAC) \cite{Craddocketal2013_cpac}. Based on our preprocessing selection, the signals underwent band-pass filtering and global signal regression. Regional mean time series were obtained via the Nilearn library \cite{Abrahametal2014}. The benchmark uses the Craddock 200 atlas \cite{Craddock2012} with 200 regions, while the controlled experiments use Harvard-Oxford with 106 retained regions. Acquisition parameters vary across the 17 sites, with the most common repetition time (TR) being $2.0$ seconds and a typical scan duration of approximately $5$--$8$ minutes per subject \cite{dimartino2014autism}.
For our analysis, we focus on the following target variables:
\begin{itemize}
    \item \texttt{DX\_GROUP}: The primary diagnostic label used for binary classification, distinguishing between individuals with ASD and Typical Controls (TC).
    \item \texttt{AGE\_AT\_SCAN}: The age of the subject at the time of the scan, in years, used for regression.
\end{itemize}

\subsubsection{Frankfurt dataset}\label{app:frankfurt_dataset}
The dataset consists of concurrent EEG-fMRI recordings from 72 healthy adults, comprising 45 females and 27 males with an age range of 19 to 48 years. Participants underwent 52-minute resting-state scans where they were instructed to remain still with their eyes closed. Despite these instructions, many participants transitioned through various stages of NREM sleep, with 64 subjects entering stage N1, 43 entering N2, and 17 reaching stage N3.
The imaging data were acquired with a repetition time ($TR$) of 2080 ms \cite{Tagliazucchietal2012}. The supplied regional time series contain 1,505 frames. Our loader removes the first five frames, leaving 1,500 frames aligned with the supplied hypnograms.
For spatial definition, we utilized a parcellation consisting of 294 regions of interest, which included 286 cortical nodes and 8 subcortical nodes \cite{Gordon2016Generation}. The benchmark uses all 294 regions. The controlled experiments average regional time series within the 15 groups specified by the supplied atlas mapping.
For each multidimensional brain time series, an associated time series indicates the sleep stage at each time point, categorized as Wake (N0), N1, N2, and N3. A distinct characteristic of this dataset is that the target variable ($y$) is also time-dependent. Consequently, when applying the rolling window approach with a specific window length and stride, the target variable must be aggregated within the same temporal window.
We construct the window-level targets separately for the two tasks:
\begin{itemize}
    \item \texttt{sleep\_frac}: The fraction of time points in the window spent asleep (N1 to N3), used for controlled regression.
    \item \texttt{sleep\_wake}: The benchmark classification label indicates whether the most frequent original sleep stage in the window is sleep (N1 to N3) or wake (N0).
\end{itemize}

\subsection{Benchmark experiments}
\subsubsection{Experiment set-up}\label{app:experiment_setup}
For each scan, we use the supplied regional mean time series from the chosen parcellation and apply a rolling-window approach: ABIDE and Frankfurt use overlapping windows of length $T$ with stride $s$, while ADHD uses the full series as one window. Within each window we compute the sample correlation matrix $\widehat C \in \mathbb{R}^{p \times p}$, where $p$ is the number of regions. These correlation matrices serve as input features for all downstream models, used both as node features and as the basis for graph construction in the GNN baselines. We report window-level test metrics, with means and standard deviations over 10 random seeds.
\paragraph{Dataset-specific settings.} Table~\ref{tab:dataset_settings} reports the parcellation, number of ROIs, window length $T$, stride, and number of subjects for each dataset.

\begin{table}[h]
\caption{Dataset-specific experimental settings. Subjects are those with the target available, split 60/20/20 by subject. For ABIDE and Frankfurt, the stride is set to $\lfloor 0.10 \cdot T \rfloor$; ABIDE uses 4 windows per subject and Frankfurt all 66. For ADHD, a single window is used, corresponding to the maximum available time-series length. Targets: age at scan (ABIDE), diagnosis (ADHD), sleep versus wake per window (Frankfurt).}
\label{tab:dataset_settings}
\centering
\begin{tabular}{llllrrr}
\toprule
Dataset & Task & Parcellation & \# ROIs & $T$ & Stride & Subjects \\
\midrule
ABIDE      & Regression     & \texttt{rois\_cc200}      & 200 & 150     & 15  & 785 \\
ADHD       & Classification & \texttt{cc200}            & 190 & full TS & N/A & 768 \\
Frankfurt  & Classification & Gordon286 + ASEG8         & 294 & 200     & 20  & 72  \\
\bottomrule
\end{tabular}
\end{table}

\paragraph{Model hyperparameters.} Table~\ref{tab:model_hyperparams} reports the most frequently selected hyperparameters for each model across benchmark runs. Hyperparameters are tuned separately for each model, dataset--task panel, and seed.

\begin{table}[h]
\caption{Most frequently selected hyperparameters across benchmark runs. All neural models were trained with $10$ seeds per dataset--task panel; GAT uses a single attention head. Graph density is $\tau =0.10$ for all graph-based benchmark models.}
\label{tab:model_hyperparams}
\centering
\begin{tabular}{lp{0.75\linewidth}}
\toprule
Model & Hyperparameters \\
\midrule
Random Forest & \raggedright \texttt{n\_estimators=400}, \texttt{max\_depth=None}, \texttt{min\_samples\_leaf=5}, \texttt{max\_features=sqrt} \tabularnewline
Ridge         & \texttt{alpha=100} \\
SVM           & \texttt{C=1}, \texttt{gamma=scale}, \texttt{kernel=rbf} \\
SVR           & \texttt{C=1}, \texttt{gamma=scale}, \texttt{kernel=rbf} \\
MLP           & \texttt{lr=1e-2}, \texttt{weight\_decay=1e-3}, \texttt{l1\_lambda=1e-4}, \texttt{hidden\_dim=64} \\
GAT (1-conv)  & \texttt{lr=1e-2}, \texttt{weight\_decay=0}, \texttt{l1\_lambda=1e-4}, \texttt{hidden\_dim=64} \\
GAT (2-conv)  & \texttt{lr=1e-2}, \texttt{weight\_decay=1e-4}, \texttt{l1\_lambda=1e-5}, \texttt{hidden\_dim=64} \\
GCN (1-conv)  & \texttt{lr=1e-2}, \texttt{weight\_decay=0}, \texttt{l1\_lambda=1e-4}, \texttt{hidden\_dim=128} \\
GCN (2-conv)  & \texttt{lr=1e-2}, \texttt{weight\_decay=1e-4}, \texttt{l1\_lambda=1e-5}, \texttt{hidden\_dim=256} \\
SAGE (1-conv) & \texttt{lr=1e-2}, \texttt{weight\_decay=0}, \texttt{l1\_lambda=0}, \texttt{hidden\_dim=256} \\
SAGE (2-conv) & \texttt{lr=1e-2}, \texttt{weight\_decay=0}, \texttt{l1\_lambda=1e-5}, \texttt{hidden\_dim=256} \\
ResidualGNN   & \raggedright \texttt{lr=1e-2}, \texttt{weight\_decay=1e-3}, \texttt{l1\_lambda=1e-4}, \texttt{hidden\_channels=128}, \texttt{hidden=128}, \texttt{num\_layers=3}, \texttt{gnn\_type=SAGEConv} \tabularnewline
\bottomrule
\end{tabular}%
\end{table}

\subsection{Controlled experiments}

We provide further details of the controlled experiments introduced in Section~\ref{sec:real_data}, followed by the full results for the progressive relaxation of assumptions, the comparisons of convolution operators and graph densities on ABIDE, and the extensions to the ADHD and Frankfurt datasets. Throughout, the independent graph $A^{\perp}_{\rm est}$ is built from the furthest non-overlapping window of the same subject, to limit the effect of autocorrelation.

\subsubsection{Progressive relaxation of assumptions}
\label{app:progressive_relaxation}

Starting from the configuration in the main text, we first introduce ReLU while retaining the vectorized readout and a fixed number of windows across $T$. The first block of Table~\ref{tab:abide_relaxation} shows that the oracle proxy remains the strongest graph construction at every $T$. Its mean accuracy exceeds that of the no-convolution baseline at $T=5,15,25$, but none of these gains reaches significance at the $5\%$ level. At $T=40$, even the oracle proxy falls below the baseline. Thus, the broad ordering persists after introducing the nonlinearity, although the evidence for a smoothing benefit is weaker. Neither estimated graph significantly improves on the baseline.

We next retain all available windows, allowing $n$ to vary with $T$. The no-convolution baseline improves at shorter window lengths: at $T=5$, its accuracy rises from $0.558$ to $0.593$. The oracle advantage is now confined to this shortest window, where it reaches $+0.037$ accuracy ($p<0.01$); at $T\geq15$, the oracle proxy no longer improves on the baseline. The remaining advantage is substantially reduced when the graph must be estimated from a separate window: at $T=5$, this replacement lowers accuracy by $0.030$ ($p<0.05$). Neither the separate-window nor the same-window graph significantly outperforms the no-convolution baseline.

Finally, replacing vectorization with mean pooling removes the remaining oracle advantage at $T=5$: the oracle-minus-baseline difference changes from $+0.037$ to $-0.003$. In this final configuration, no graph construction significantly improves on the corresponding no-convolution baseline at any tested $T$. For ABIDE, the benefit observed under the initial restrictions therefore does not survive the full progression toward the deployed pipeline, connecting the controlled experiments to the benchmark findings.

\begin{table}[htbp]
\caption{\textbf{Progressive relaxation of the assumptions of Table~\ref{tab:abide_scenarios}.} Same cell and conventions, $\mathrm{SageMean}$, $\tau=0.10$; each block adds one relaxation to the previous one, starting from Table~\ref{tab:abide_scenarios}. Mean $\pm$ 1 SD over 20 seeds; paired $t$-test over seeds against $\widehat C$ (left) or zero (right): $^{*}p<0.10$, $^{**}p<0.05$, $^{***}p<0.01$.}
\label{tab:abide_relaxation}
\centering
\footnotesize
\setlength{\tabcolsep}{0pt}
\begin{tabular*}{\linewidth}{@{\extracolsep{\fill}} c ccccc cccc}
\toprule
$T$ & $\widehat C$ & $A^{\text{oracle}}_{\rm est}$ & $A^\perp_{\rm est}$ & $\widehat A$ & $A^{\mathrm{pop}}$ & $A^{\text{oracle}}_{\rm est}-\widehat C$ & $A^\perp_{\rm est}-A^{\text{oracle}}_{\rm est}$ & $\widehat A-A^\perp_{\rm est}$ & $A^{\mathrm{pop}}-\widehat C$ \\
\midrule
\multicolumn{10}{l}{ReLU activation (vectorized readout, $n$ fixed)} \\
5 & \msd{0.558}{}{.037} & \msd{0.567}{}{.044} & \msd{0.537}{**}{.037} & \msd{0.546}{}{.038} & \msd{0.550}{}{.031} & \gs{+0.009}{} & \gs{-0.029}{***} & \gs{+0.009}{} & \gs{-0.008}{} \\
15 & \msd{0.617}{}{.051} & \msd{0.632}{}{.044} & \msd{0.615}{}{.051} & \msd{0.620}{}{.045} & \msd{0.609}{}{.051} & \gs{+0.015}{} & \gs{-0.017}{*} & \gs{+0.005}{} & \gs{-0.009}{} \\
25 & \msd{0.643}{}{.045} & \msd{0.660}{*}{.032} & \msd{0.641}{}{.048} & \msd{0.653}{}{.030} & \msd{0.640}{}{.046} & \gs{+0.017}{*} & \gs{-0.019}{*} & \gs{+0.012}{} & \gs{-0.003}{} \\
40 & \msd{0.680}{}{.035} & \msd{0.676}{}{.031} & \msd{0.666}{**}{.033} & \msd{0.662}{***}{.031} & \msd{0.656}{**}{.046} & \gs{-0.005}{} & \gs{-0.009}{**} & \gs{-0.004}{} & \gs{-0.024}{**} \\
\midrule
\multicolumn{10}{l}{ReLU activation and $n$ free (vectorized readout)} \\
5 & \msd{0.593}{}{.057} & \msd{0.630}{***}{.044} & \msd{0.600}{}{.064} & \msd{0.601}{}{.056} & \msd{0.601}{}{.059} & \gs{+0.037}{***} & \gs{-0.030}{**} & \gs{+0.002}{} & \gs{+0.008}{} \\
15 & \msd{0.669}{}{.036} & \msd{0.667}{}{.034} & \msd{0.663}{}{.034} & \msd{0.651}{**}{.039} & \msd{0.653}{***}{.033} & \gs{-0.002}{} & \gs{-0.004}{} & \gs{-0.013}{} & \gs{-0.016}{***} \\
25 & \msd{0.681}{}{.036} & \msd{0.680}{}{.040} & \msd{0.673}{}{.039} & \msd{0.676}{}{.041} & \msd{0.682}{}{.037} & \gs{-0.001}{} & \gs{-0.007}{} & \gs{+0.003}{} & \gs{+0.001}{} \\
40 & \msd{0.676}{}{.031} & \msd{0.672}{}{.036} & \msd{0.670}{}{.036} & \msd{0.671}{}{.040} & \msd{0.667}{}{.035} & \gs{-0.004}{} & \gs{-0.003}{} & \gs{+0.002}{} & \gs{-0.009}{} \\
\midrule
\multicolumn{10}{l}{ReLU activation, $n$ free and mean pooling (deployed pipeline)} \\
5 & \msd{0.623}{}{.038} & \msd{0.620}{}{.049} & \msd{0.627}{}{.050} & \msd{0.626}{}{.051} & \msd{0.627}{}{.045} & \gs{-0.003}{} & \gs{+0.007}{} & \gs{-0.001}{} & \gs{+0.004}{} \\
15 & \msd{0.672}{}{.038} & \msd{0.657}{**}{.042} & \msd{0.652}{***}{.042} & \msd{0.659}{**}{.042} & \msd{0.657}{**}{.042} & \gs{-0.015}{**} & \gs{-0.005}{} & \gs{+0.007}{**} & \gs{-0.015}{**} \\
25 & \msd{0.666}{}{.032} & \msd{0.663}{}{.032} & \msd{0.664}{}{.033} & \msd{0.667}{}{.033} & \msd{0.666}{}{.030} & \gs{-0.003}{} & \gs{+0.000}{} & \gs{+0.003}{} & \gs{+0.000}{} \\
40 & \msd{0.668}{}{.041} & \msd{0.664}{}{.035} & \msd{0.658}{*}{.033} & \msd{0.660}{}{.027} & \msd{0.661}{**}{.032} & \gs{-0.004}{} & \gs{-0.007}{} & \gs{+0.003}{} & \gs{-0.007}{**} \\
\bottomrule
\end{tabular*}
\end{table}

\subsubsection{Extensions to other convolutions and datasets}

GAT lies outside our linear theory, because its attention weights are computed from the same noisy matrix as the features and therefore inherit its noise. GAT follows the same qualitative pattern as SageMean: in the simulation-like setting, the oracle proxy helps at short windows, but neither estimated graph significantly outperforms its matched no-convolution baseline. In the deployed setting, neither operator benefits significantly from any graph construction.

\begin{table}[htbp]
\caption{\textbf{GAT in place of $\mathrm{SageMean}$.} Same cell and conventions as Table~\ref{tab:abide_scenarios}, $\tau=0.10$, with one GAT convolution (8 heads of 8); the $\widehat C$ column is the same architecture with message passing removed. Mean $\pm$ 1 SD over 20 seeds; paired $t$-test over seeds against $\widehat C$ (left) or zero (right): $^{*}p<0.10$, $^{**}p<0.05$, $^{***}p<0.01$.}
\label{tab:abide_gat}
\centering
\footnotesize
\setlength{\tabcolsep}{0pt}
\begin{tabular*}{\linewidth}{@{\extracolsep{\fill}} c ccccc cccc}
\toprule
$T$ & $\widehat C$ & $A^{\text{oracle}}_{\rm est}$ & $A^\perp_{\rm est}$ & $\widehat A$ & $A^{\mathrm{pop}}$ & $A^{\text{oracle}}_{\rm est}-\widehat C$ & $A^\perp_{\rm est}-A^{\text{oracle}}_{\rm est}$ & $\widehat A-A^\perp_{\rm est}$ & $A^{\mathrm{pop}}-\widehat C$ \\
\midrule
\multicolumn{10}{l}{Simulation-like scenario (identity activation, vectorized readout, $n$ fixed)} \\
5 & \msd{0.558}{}{.033} & \msd{0.585}{***}{.031} & \msd{0.547}{}{.037} & \msd{0.557}{}{.031} & \msd{0.554}{}{.034} & \gs{+0.027}{***} & \gs{-0.038}{***} & \gs{+0.010}{} & \gs{-0.004}{} \\
15 & \msd{0.621}{}{.040} & \msd{0.643}{***}{.035} & \msd{0.608}{*}{.032} & \msd{0.618}{}{.038} & \msd{0.605}{***}{.032} & \gs{+0.022}{***} & \gs{-0.035}{***} & \gs{+0.010}{} & \gs{-0.016}{***} \\
25 & \msd{0.649}{}{.037} & \msd{0.654}{}{.033} & \msd{0.637}{*}{.035} & \msd{0.643}{}{.036} & \msd{0.638}{**}{.032} & \gs{+0.005}{} & \gs{-0.017}{**} & \gs{+0.006}{} & \gs{-0.011}{**} \\
40 & \msd{0.680}{}{.047} & \msd{0.663}{**}{.037} & \msd{0.662}{***}{.035} & \msd{0.665}{***}{.038} & \msd{0.654}{***}{.035} & \gs{-0.018}{**} & \gs{-0.001}{} & \gs{+0.003}{} & \gs{-0.027}{***} \\
\midrule
\multicolumn{10}{l}{Deployed pipeline (ReLU activation, mean pooling, $n$ free)} \\
5 & \msd{0.633}{}{.049} & \msd{0.616}{**}{.036} & \msd{0.620}{*}{.037} & \msd{0.631}{}{.048} & \msd{0.615}{**}{.042} & \gs{-0.017}{**} & \gs{+0.004}{} & \gs{+0.011}{} & \gs{-0.018}{**} \\
15 & \msd{0.677}{}{.036} & \msd{0.665}{**}{.033} & \msd{0.662}{**}{.031} & \msd{0.664}{**}{.036} & \msd{0.667}{*}{.035} & \gs{-0.012}{**} & \gs{-0.003}{} & \gs{+0.003}{} & \gs{-0.010}{*} \\
25 & \msd{0.670}{}{.024} & \msd{0.658}{**}{.026} & \msd{0.662}{*}{.023} & \msd{0.663}{**}{.024} & \msd{0.668}{}{.026} & \gs{-0.012}{**} & \gs{+0.004}{} & \gs{+0.001}{} & \gs{-0.002}{} \\
40 & \msd{0.670}{}{.031} & \msd{0.659}{**}{.031} & \msd{0.659}{**}{.028} & \msd{0.660}{**}{.035} & \msd{0.661}{*}{.031} & \gs{-0.011}{**} & \gs{+0.000}{} & \gs{+0.000}{} & \gs{-0.009}{*} \\
\bottomrule
\end{tabular*}
\end{table}

On ADHD, the oracle proxy provides a modest benefit at $T=7$, which is lost when the graph is estimated from a separate window; at $T=40$, even the oracle proxy falls below the baseline. In the deployed setting, no subject-specific graph significantly improves prediction, although the population graph provides a small gain at $T=40$.

\begin{table}[htbp]
\caption{\textbf{ADHD.} Subject-level test $R^2$ for the inattention score on ADHD-200 (CC200, $p=190$) with $\mathrm{SageMean}(A,\widehat C)$, one convolution, $\tau=0.10$, in the simulation-like scenario and the deployed pipeline. Mean $\pm$ 1 SD over 20 seeds; paired $t$-test over seeds against $\widehat C$ (left) or zero (right): $^{*}p<0.10$, $^{**}p<0.05$, $^{***}p<0.01$. The train-mean predictor has $R^2=-0.03$.}
\label{tab:adhd_scenarios}
\centering
\footnotesize
\setlength{\tabcolsep}{0pt}
\begin{tabular*}{\linewidth}{@{\extracolsep{\fill}} c ccccc cccc}
\toprule
$T$ & $\widehat C$ & $A^{\text{oracle}}_{\rm est}$ & $A^\perp_{\rm est}$ & $\widehat A$ & $A^{\mathrm{pop}}$ & $A^{\text{oracle}}_{\rm est}-\widehat C$ & $A^\perp_{\rm est}-A^{\text{oracle}}_{\rm est}$ & $\widehat A-A^\perp_{\rm est}$ & $A^{\mathrm{pop}}-\widehat C$ \\
\midrule
\multicolumn{10}{l}{Simulation-like scenario (identity activation, vectorized readout, $n$ fixed)} \\
5 & \msd{0.243}{}{.060} & \msd{0.261}{}{.063} & \msd{0.219}{*}{.069} & \msd{0.225}{}{.063} & \msd{0.251}{}{.057} & \gs{+0.018}{} & \gs{-0.042}{***} & \gs{+0.006}{} & \gs{+0.008}{} \\
7 & \msd{0.318}{}{.047} & \msd{0.333}{**}{.057} & \msd{0.267}{***}{.061} & \msd{0.312}{}{.058} & \msd{0.320}{}{.055} & \gs{+0.015}{**} & \gs{-0.066}{***} & \gs{+0.045}{***} & \gs{+0.002}{} \\
15 & \msd{0.463}{}{.043} & \msd{0.469}{}{.057} & \msd{0.443}{**}{.051} & \msd{0.448}{*}{.053} & \msd{0.466}{}{.059} & \gs{+0.005}{} & \gs{-0.026}{***} & \gs{+0.006}{} & \gs{+0.002}{} \\
25 & \msd{0.531}{}{.035} & \msd{0.526}{}{.043} & \msd{0.511}{***}{.043} & \msd{0.516}{**}{.040} & \msd{0.532}{}{.046} & \gs{-0.006}{} & \gs{-0.015}{} & \gs{+0.005}{} & \gs{+0.001}{} \\
40 & \msd{0.543}{}{.063} & \msd{0.523}{***}{.058} & \msd{0.526}{*}{.051} & \msd{0.525}{**}{.054} & \msd{0.529}{*}{.050} & \gs{-0.020}{***} & \gs{+0.003}{} & \gs{-0.002}{} & \gs{-0.014}{*} \\
\midrule
\multicolumn{10}{l}{Deployed pipeline (ReLU activation, mean pooling, $n$ free)} \\
5 & \msd{0.289}{}{.027} & \msd{0.276}{**}{.027} & \msd{0.280}{***}{.025} & \msd{0.273}{***}{.027} & \msd{0.280}{*}{.026} & \gs{-0.013}{**} & \gs{+0.003}{} & \gs{-0.007}{} & \gs{-0.009}{*} \\
7 & \msd{0.329}{}{.032} & \msd{0.319}{*}{.033} & \msd{0.328}{}{.033} & \msd{0.314}{***}{.031} & \msd{0.331}{}{.032} & \gs{-0.009}{*} & \gs{+0.008}{*} & \gs{-0.014}{***} & \gs{+0.002}{} \\
15 & \msd{0.411}{}{.036} & \msd{0.398}{**}{.044} & \msd{0.401}{*}{.043} & \msd{0.382}{***}{.044} & \msd{0.416}{}{.040} & \gs{-0.014}{**} & \gs{+0.004}{} & \gs{-0.020}{***} & \gs{+0.005}{} \\
25 & \msd{0.450}{}{.055} & \msd{0.425}{***}{.054} & \msd{0.424}{**}{.061} & \msd{0.423}{***}{.054} & \msd{0.444}{}{.055} & \gs{-0.025}{***} & \gs{-0.001}{} & \gs{-0.001}{} & \gs{-0.006}{} \\
40 & \msd{0.432}{}{.062} & \msd{0.428}{}{.050} & \msd{0.411}{}{.074} & \msd{0.421}{}{.046} & \msd{0.452}{**}{.050} & \gs{-0.004}{} & \gs{-0.018}{} & \gs{+0.011}{} & \gs{+0.019}{**} \\
\bottomrule
\end{tabular*}
\end{table}

Frankfurt differs because sleep state changes within a scan: changing $T$ changes both connectivity-estimation noise and the window-level sleep fraction being predicted. The full-series graph also averages across sleep states and may therefore poorly represent an individual window. In the simulation-like setting, the oracle proxy is strongest at the shortest windows but weakest at $T=45$, while the separate-window graph performs best from $T=14$ onward. Graph benefits persist in the deployed setting, with all four constructions significantly improving prediction at $T=25$ and $45$.

\begin{table}[htbp]
\caption{\textbf{Frankfurt.} Window-level test $R^2$ for the fraction of the window spent asleep on the Frankfurt dataset (15 networks) with $\mathrm{SageMean}(A,\widehat C)$, one convolution, $\tau=0.10$, in the simulation-like scenario and the deployed pipeline. The target is defined per window, so $A^{\text{oracle}}_{\rm est}$ is built from a series whose label changes within the scan. Mean $\pm$ 1 SD over 20 seeds; paired $t$-test over seeds against $\widehat C$ (left) or zero (right): $^{*}p<0.10$, $^{**}p<0.05$, $^{***}p<0.01$. The train-mean predictor has $R^2=-0.03$.}
\label{tab:frankfurt_scenarios}
\centering
\footnotesize
\setlength{\tabcolsep}{0pt}
\begin{tabular*}{\linewidth}{@{\extracolsep{\fill}} c ccccc cccc}
\toprule
$T$ & $\widehat C$ & $A^{\text{oracle}}_{\rm est}$ & $A^\perp_{\rm est}$ & $\widehat A$ & $A^{\mathrm{pop}}$ & $A^{\text{oracle}}_{\rm est}-\widehat C$ & $A^\perp_{\rm est}-A^{\text{oracle}}_{\rm est}$ & $\widehat A-A^\perp_{\rm est}$ & $A^{\mathrm{pop}}-\widehat C$ \\
\midrule
\multicolumn{10}{l}{Simulation-like scenario (identity activation, vectorized readout, $n$ fixed)} \\
5 & \msd{0.069}{}{.067} & \msd{0.126}{***}{.090} & \msd{0.066}{}{.072} & \msd{0.060}{***}{.068} & \msd{0.066}{}{.071} & \gs{+0.057}{***} & \gs{-0.060}{***} & \gs{-0.007}{} & \gs{-0.003}{} \\
8 & \msd{0.141}{}{.052} & \msd{0.193}{***}{.070} & \msd{0.145}{}{.059} & \msd{0.144}{}{.056} & \msd{0.145}{}{.047} & \gs{+0.052}{***} & \gs{-0.048}{***} & \gs{-0.002}{} & \gs{+0.005}{} \\
14 & \msd{0.246}{}{.073} & \msd{0.262}{}{.082} & \msd{0.271}{***}{.070} & \msd{0.252}{**}{.073} & \msd{0.249}{}{.073} & \gs{+0.016}{} & \gs{+0.009}{} & \gs{-0.018}{***} & \gs{+0.004}{} \\
25 & \msd{0.374}{}{.063} & \msd{0.369}{}{.085} & \msd{0.398}{***}{.066} & \msd{0.381}{*}{.064} & \msd{0.379}{**}{.064} & \gs{-0.004}{} & \gs{+0.029}{**} & \gs{-0.017}{***} & \gs{+0.005}{**} \\
45 & \msd{0.461}{}{.065} & \msd{0.445}{}{.083} & \msd{0.488}{***}{.076} & \msd{0.461}{}{.071} & \msd{0.477}{***}{.066} & \gs{-0.016}{} & \gs{+0.044}{***} & \gs{-0.028}{***} & \gs{+0.016}{***} \\
\midrule
\multicolumn{10}{l}{Deployed pipeline (ReLU activation, mean pooling, $n$ free)} \\
5 & \msd{0.051}{}{.049} & \msd{0.063}{***}{.054} & \msd{0.054}{}{.058} & \msd{0.058}{**}{.051} & \msd{0.058}{**}{.049} & \gs{+0.012}{***} & \gs{-0.009}{***} & \gs{+0.004}{} & \gs{+0.007}{**} \\
8 & \msd{0.096}{}{.053} & \msd{0.104}{***}{.054} & \msd{0.098}{}{.057} & \msd{0.097}{}{.059} & \msd{0.097}{}{.058} & \gs{+0.009}{***} & \gs{-0.006}{**} & \gs{-0.001}{} & \gs{+0.002}{} \\
14 & \msd{0.165}{}{.047} & \msd{0.181}{***}{.061} & \msd{0.178}{**}{.068} & \msd{0.174}{*}{.061} & \msd{0.171}{}{.064} & \gs{+0.016}{***} & \gs{-0.003}{} & \gs{-0.004}{} & \gs{+0.006}{} \\
25 & \msd{0.246}{}{.063} & \msd{0.276}{***}{.062} & \msd{0.283}{***}{.067} & \msd{0.272}{***}{.059} & \msd{0.270}{***}{.059} & \gs{+0.029}{***} & \gs{+0.007}{**} & \gs{-0.011}{***} & \gs{+0.024}{***} \\
45 & \msd{0.316}{}{.063} & \msd{0.344}{***}{.064} & \msd{0.356}{***}{.066} & \msd{0.347}{***}{.060} & \msd{0.340}{***}{.063} & \gs{+0.029}{***} & \gs{+0.012}{***} & \gs{-0.009}{**} & \gs{+0.025}{***} \\
\bottomrule
\end{tabular*}
\end{table}

\subsubsection{Effect of graph density}

We repeat the ABIDE simulation-like experiment at graph densities $\tau=0.05$ and $0.20$, comparing them with $\tau=0.10$ in the main text. At $T=5$, the oracle proxy significantly improves accuracy at all three densities, with gains of $0.019$, $0.022$, and $0.024$ for $\tau=0.05$, $0.10$, and $0.20$, respectively. However, neither the separate-window nor the same-window graph significantly outperforms the no-convolution baseline at any tested density or $T$. The oracle benefit also varies with density: at $T=25$, it is significant only for $\tau=0.10$, while at $T=40$ its mean performance falls below the baseline at every density. Thus, making the graph sparser or denser does not consistently recover a predictive advantage in this setting.

\begin{table}[htbp]
\caption{\textbf{Threshold ablation of Table~\ref{tab:abide_scenarios}.} Same cell and conventions, $\mathrm{SageMean}$, simulation-like scenario, at $\tau=0.05$ and $\tau=0.20$ ($\tau=0.10$ is Table~\ref{tab:abide_scenarios}); the $\widehat C$ column does not depend on $\tau$. Mean $\pm$ 1 SD over 20 seeds; paired $t$-test over seeds against $\widehat C$ (left) or zero (right): $^{*}p<0.10$, $^{**}p<0.05$, $^{***}p<0.01$.}
\label{tab:abide_tau}
\centering
\footnotesize
\setlength{\tabcolsep}{0pt}
\begin{tabular*}{\linewidth}{@{\extracolsep{\fill}} c ccccc cccc}
\toprule
$T$ & $\widehat C$ & $A^{\text{oracle}}_{\rm est}$ & $A^\perp_{\rm est}$ & $\widehat A$ & $A^{\mathrm{pop}}$ & $A^{\text{oracle}}_{\rm est}-\widehat C$ & $A^\perp_{\rm est}-A^{\text{oracle}}_{\rm est}$ & $\widehat A-A^\perp_{\rm est}$ & $A^{\mathrm{pop}}-\widehat C$ \\
\midrule
\multicolumn{10}{l}{$\tau=0.05$} \\
5 & \msd{0.560}{}{.033} & \msd{0.580}{**}{.034} & \msd{0.563}{}{.033} & \msd{0.563}{}{.033} & \msd{0.567}{}{.035} & \gs{+0.019}{**} & \gs{-0.016}{**} & \gs{-0.000}{} & \gs{+0.006}{} \\
15 & \msd{0.625}{}{.040} & \msd{0.628}{}{.035} & \msd{0.616}{}{.039} & \msd{0.616}{}{.039} & \msd{0.623}{}{.040} & \gs{+0.003}{} & \gs{-0.013}{**} & \gs{+0.001}{} & \gs{-0.003}{} \\
25 & \msd{0.651}{}{.037} & \msd{0.654}{}{.043} & \msd{0.649}{}{.051} & \msd{0.651}{}{.045} & \msd{0.652}{}{.045} & \gs{+0.003}{} & \gs{-0.005}{} & \gs{+0.002}{} & \gs{+0.001}{} \\
40 & \msd{0.682}{}{.034} & \msd{0.661}{**}{.030} & \msd{0.668}{**}{.033} & \msd{0.660}{**}{.035} & \msd{0.668}{}{.035} & \gs{-0.021}{**} & \gs{+0.007}{} & \gs{-0.008}{} & \gs{-0.014}{} \\
\midrule
\multicolumn{10}{l}{$\tau=0.20$} \\
5 & \msd{0.560}{}{.033} & \msd{0.584}{***}{.036} & \msd{0.566}{}{.033} & \msd{0.558}{}{.037} & \msd{0.567}{}{.038} & \gs{+0.024}{***} & \gs{-0.019}{***} & \gs{-0.008}{} & \gs{+0.006}{} \\
15 & \msd{0.625}{}{.040} & \msd{0.629}{}{.033} & \msd{0.622}{}{.041} & \msd{0.623}{}{.037} & \msd{0.617}{}{.034} & \gs{+0.004}{} & \gs{-0.008}{*} & \gs{+0.001}{} & \gs{-0.008}{} \\
25 & \msd{0.651}{}{.037} & \msd{0.661}{}{.040} & \msd{0.644}{}{.049} & \msd{0.651}{}{.045} & \msd{0.653}{}{.035} & \gs{+0.009}{} & \gs{-0.017}{**} & \gs{+0.007}{} & \gs{+0.002}{} \\
40 & \msd{0.682}{}{.034} & \msd{0.676}{}{.031} & \msd{0.671}{*}{.034} & \msd{0.664}{**}{.036} & \msd{0.675}{}{.032} & \gs{-0.006}{} & \gs{-0.005}{} & \gs{-0.008}{} & \gs{-0.007}{} \\
\bottomrule
\end{tabular*}
\end{table}

\subsection{Practical recommendations}\label{app:recommendations}
Our results suggest a few guidelines for connectome GNN pipelines. First, convolutions help only at short windows: when longer recordings are available, longer windows combined with graph-agnostic models are likely to perform better, so convolutions should be used with care. Second, estimating the graph from a longer series than the window, as the oracle proxy does, extends the range of $T$ over which convolutions help, provided the target does not change within the scan. Third, building the graph from the same window as the features adds a double-dipping term of no consistent sign, which makes results less stable rather than systematically worse; cross-fitting removes this term but not the graph-estimation error, which is the main bottleneck. Fourth, smoothing can only help when the target is community-smooth: if the signal sits in single edges or in contrasts within a community, smoothing removes it. Finally, when these conditions are met, the graph density must be tuned; otherwise, no choice of density recovers a gain.

\section{Compute resources}\label{app:compute}

All experiments were run on CPU nodes equipped with Intel Broadwell (28 cores at 2.4 GHz, 64 GB memory) and Intel Skylake (40 cores at 2.4 GHz, 96 GB memory) processors, for a total of approximately $3{,}000$ CPU hours.

\section{Broader impacts}\label{app:broader_impacts}

Our work is a theoretical and methodological analysis of when graph neural network methods provide statistical benefits over simpler baselines in brain connectome analysis. By clarifying the regimes in which these methods help versus hurt, we hope to inform more reliable use of machine learning in clinical neuroscience applications such as the diagnosis of psychiatric and neurological disorders. We do not introduce new models, datasets, or deployment ready systems, and we do not foresee direct negative societal impacts of this work.

\section{LLM usage declaration}\label{app:llm_declaration}

We used large language models (LLMs) as coding assistants to help implement and run some of the experimental pipelines, and as writing assistants to copy edit and improve the clarity of the manuscript.

\end{document}